\documentclass[a4paper, nobind, 12pt, hidelinks]{OxfordTeXThesis} 

\backgroundsetup{contents={}} 

\correctionstrue

\usepackage[style=numeric-comp, sorting=none, backend=biber, doi=true, isbn=false]{biblatex}
\newcommand*{\bibtitle}{References}

\usepackage[style=alttree, nonumberlist, toc, acronym]{glossaries-extra}
\makenoidxglossaries 

\newacronym{AEMO}{AEMO}{Australian Energy Market Operator}
\newacronym{BESS}{BESS}{Battery Energy Storage System}
\newacronym{NEM}{NEM}{National Electricity Market}
\newacronym{SOC}{SOC}{State of Charge}
\newacronym{LP}{LP}{Linear Programming}
\newacronym{MILP}{MILP}{Mixed-Integer Linear Program}
\newacronym{NSW}{NSW}{New South Wales}
\newacronym{QLD}{QLD}{Queensland}
\newacronym{SA}{SA}{South Australia}
\newacronym{VIC}{VIC}{Victoria}
\newacronym{TAS}{TAS}{Tasmania}
\newacronym{CAPEX}{CAPEX}{Capital Expenditure}
\newacronym{FCAS}{FCAS}{Frequency Control Ancillary Services}
\newacronym{NER}{NER}{National Electricity Rules}
\newacronym{ML}{ML}{Machine Learning}
\newacronym{OPEX}{OPEX}{Operational Expenditure}
\newacronym{LCOS}{LCOS}{Levelised Cost of Storage}
\newacronym{WACC}{WACC}{Weighted-Average Cost of Capital}
\newacronym{NPV}{NPV}{Net Present Value}
\newacronym{IRR}{IRR}{Internal Rate of Return}

\title{Optimising Battery Energy Storage System Trading via Energy Market Operator Price Forecast}
\author{Aymeric Marie Fran\c{c}ois Fabre}
\college{Department of Engineering Science}

\degree{Master of Science}
\degreedate{Trinity term, 2025}
\supervisors{Professor Thomas Morstyn}

\DeclareSIUnit\bar{bar} 
\DeclareSIUnit\mrad{mrad}
\DeclareSIUnit{\wtpercent}{wt\%}

\DeclareGraphicsExtensions{.png, .pdf} 

\begin{document}



\setlength{\textbaselineskip}{22pt plus2pt} 


\setlength{\frontmatterbaselineskip}{22pt plus2pt} 

\setlength{\baselineskip}{\textbaselineskip}



\setcounter{secnumdepth}{2}
\setcounter{tocdepth}{2}
\setcounter{minitocdepth}{2}



\begin{romanpages} 



\maketitle


\begin{abstract}

In electricity markets around the world, the ability to anticipate price movements with precision can be the difference between profit and loss, especially for fast-acting assets like battery energy storage systems (BESS). As grid volatility increases due to renewables and market decentralisation, operators and forecasters alike face growing pressure to transform prediction into strategy. Yet while forecast data is abundant, especially in advanced markets like Australia’s National Electricity Market (NEM), its practical value in driving real-world BESS trading decisions remains largely unexplored. This thesis dives into that gap.
\\\\
This work addresses a key research question: Can the accuracy of the Australian Energy Market Operator (AEMO) energy price forecasts be systematically leveraged to develop a reliable and profitable battery energy storage system trading algorithm? Despite the availability of AEMO price forecasts, no existing framework evaluates their reliability or incorporates them into practical BESS trading strategies. By analysing patterns in forecast accuracy based on time of day, forecast horizon, and regional variations, this project creates a novel, forecast-informed BESS trading model to optimise arbitrage financial returns. The performance of this forecast-driven algorithm is benchmarked against a basic trading algorithm with no knowledge of forecast data. The study further explores the potential of machine learning techniques to predict future energy prices by enhancing AEMO forecasts to govern a more advanced trading strategy. The research outcomes will inform future improvements in energy market trading models and promote more efficient BESS integration into market operations.

\end{abstract}
\begin{acknowledgements}
 	First and foremost, I would like to express my sincere gratitude to my supervisor, Professor Thomas Morstyn, and his doctoral student, Joseph Cary, for their invaluable guidance, support, and encouragement throughout the course of this thesis. Their expertise and insights were instrumental in shaping the direction and quality of my research.
\\\\
I am grateful to the Australian Energy Market Operator for providing open-source energy market data, without which this thesis would not have been possible.
\\\\
To my parents and brothers, thank you for your unwavering support, love, and belief in me, not just during this academic journey, but throughout my life. Your encouragement has been a constant source of strength.
\\\\
I am deeply thankful to God for His infinite support and guidance through the many ups and downs. Faith has been an anchor during the most challenging moments.
\\\\
Finally, I would like to thank myself. For working hard to afford this degree, for pushing through the long days and nights, and for never giving up, this thesis is a testament to that persistence.
\end{acknowledgements}

\dominitoc 

\flushbottom

\tableofcontents





\renewcommand{\glossarypreamble}{\glsfindwidesttoplevelname[\currentglossary]}
\printnoidxglossary[title=Abbreviations, toctitle=Abbreviations, type=acronym, nogroupskip]


\end{romanpages}


\begin{savequote}[\textwidth/2]
When shall we three meet again
in thunder, lightning, or in rain?
\\
When the hurlyburly’s done,
when the battle’s lost and won.
\qauthor{\textbf{---}Shakespeare, \textit{Macbeth}}
\end{savequote}

\chapter{Introduction}
\label{ch: Introduction}

\adjustmtc
\minitoc


\section{Background}
As the global energy transition accelerates, Battery Energy Storage Systems (\gls{BESS}) are becoming indispensable to grid stability, decarbonisation, and market optimisation. Yet, the full potential of BESS remains untapped, not because of technological limitations, but due to operational strategies that fail to exploit market dynamics effectively. This study tackles a central challenge: how can we trade electricity with a BESS more intelligently by leveraging forecast data, and does it actually make a measurable difference? Focusing on the Australian National Electricity Market (\gls{NEM}), which is known for its price volatility, decentralised design, and advanced forecasting practices, this research serves as a testbed for evaluating the true economic value of forecast-informed BESS trading. The implications extend far beyond Australia. Lessons drawn here are directly applicable to deregulated markets like those in the UK and USA, and to emerging markets attempting to integrate intermittent renewables with limited infrastructure. This work is intended for BESS owners seeking revenue maximisation, system operators responsible for balancing the grid, and forecasters developing models that must translate into real-world decision-making value. By evaluating how forecasts impact BESS behaviour and profitability, the study provides insight not only into energy market optimisation, but also offers broader methodological relevance for forecast evaluation across industries such as finance, agriculture, and supply chain logistics. This is a practical investigation into the difference between predictive power and actionable intelligence and why that difference matters.

\section{Literature Review}
The optimisation of BESS trading strategies within wholesale electricity markets has garnered increasing interest due to the critical role that energy storage plays in enhancing grid stability and integrating renewable energy sources, with BESS expected to become an increasingly central component of the energy system as outlined in the Australian Energy Market Operator's (\gls{AEMO}) latest Integrated System Plan \cite{AEMO_ISP, BESS_stability_renewable_integration}. Given Australia's rapid deployment of renewable energy, unique market structure, and the availability of granular, publicly accessible forecast and pricing data through AEMO, the NEM provides a particularly compelling context in which to explore the development of forecast-informed BESS trading strategies. While it is highly likely that private entities have developed proprietary trading algorithms incorporating AEMO forecasts to enhance price predictions \cite{probabilistic_NEM_forecast, proprietary_algorithms, LSTM_forecasting}, there is a conspicuous lack of publicly available research evaluating the reliability of these forecasts or systematically leveraging them to inform BESS trading strategies. This absence of transparent, peer-reviewed studies represents a significant gap in the literature and hinders the development of best practices that could improve the efficiency of energy storage integration across the NEM.

\vspace{20pt}

The role of BESS in energy markets is multifaceted, with applications ranging from peak shaving and load shifting to frequency regulation and arbitrage trading \cite{BESS_applications, role_of_BESS2, Revisiting_Energy_Storage_BCG}. Peak shaving and load shifting manage demand by storing energy during low-use periods and discharging during peaks. Frequency regulation stabilises grid frequency through rapid response, whilst arbitrage trading exploits price fluctuations to generate profit. It is the latter of these applications, arbitrage, on which this study is focused. Although BESS technologies have relatively high embodied carbon due to materials and manufacturing processes \cite{BESS_LCA}, their ability to displace fossil fuel generation by shifting renewable energy to periods of peak demand can lead to a net reduction in carbon emissions across the energy system \cite{BESS_CO2_reduction}. By storing excess electricity during periods of low demand and discharging during peak demand, BESS not only enhances grid resilience but also provides financial incentives for market participants, especially solar and wind farms looking to invest in on-site storage \cite{hybrid_projects_for_the_win, hybrid_projects_for_the_wind}. The viability of arbitrage strategies, however, is highly dependent on price volatility and the ability to anticipate price fluctuations with a reasonable degree of confidence \cite{Arbitrage_Under_Price_Uncertainty, Arbitrage_Under_Price_Uncertainty2}. Given that price spikes and troughs can be short-lived, a well-optimised trading strategy requires not only an understanding of current market conditions but also a reliable forecast of future prices.
\\\\
Energy market price forecasting is a well-established field, typically relying on a combination of historical price trends, demand-supply modelling, and external factors such as weather conditions and fuel prices \cite{electricity_price_forecasting_review, data_driven_energy_price_forecasting_review}. The AEMO provides energy price forecasts for different regions and time horizons \cite{AEMO_Dashboard}, which offers market participants the ability to make informed decisions regarding energy trading and dispatch strategies \cite{FANTASTIC_AEMO_FORECAST_PAPER}. However, the publication of price forecasts by a system operator such as AEMO may itself influence market behaviour, potentially creating feedback effects as participants adjust their actions in anticipation of others responding to the forecast. Furthermore, the accuracy of these forecasts is inherently uncertain, with variations arising from factors such as forecast horizon length, market volatility, and regional disparities. Despite the widespread availability of AEMO’s forecasts, no standard framework exists for quantifying their reliability or assessing their potential as a driver for BESS trading decision-making. This research aims to fill that void by conducting a systematic evaluation of forecast accuracy and developing trading algorithms that adapt dynamically to observed forecast performance.
\\\\
Existing BESS trading strategies typically fall into two broad categories: rule-based heuristics and optimisation frameworks that incorporate predictive analytics as inputs to decision-making \cite{using_LP_optimisation_in_BESS}. Rule-based approaches rely on predefined thresholds and decision rules, such as charging when prices fall below a certain threshold and discharging when they exceed a target level \cite{rule_based_trading}. While these strategies are simple to implement, they lack adaptability and often fail to respond effectively to the stochastic nature of energy prices. In contrast, more advanced approaches use statistical modelling or machine learning (\gls{ML}) to generate short-term price forecasts, which then inform optimisation-based trading strategies designed to maximise arbitrage revenues or meet specific operational constraints. Whilst \cite{FANTASTIC_AEMO_FORECAST_PAPER} utilised AEMO data to compare revenue outcomes under perfect foresight versus those obtained using AEMO’s forecast to quantify theoretical opportunity loss due to forecast imperfections, the present work adopts a different emphasis. Instead of comparing forecasts to perfect information, this study compares the AEMO forecast to a zero-foresight baseline to directly assess its added value in operational contexts. In addition, forecast accuracy is evaluated across multiple temporal scales to understand how lead time affects predictive performance. Unlike \cite{FANTASTIC_AEMO_FORECAST_PAPER}, this work also integrates AEMO forecasts into a machine learning model to enhance forecast accuracy and explores how these improved forecasts translate into better trading outcomes. Finally, a financial analysis is conducted to quantify the realised economic benefits of forecast-aware versus naïve strategies to offer practical insights for BESS operators, market participants, and system planners.
\\\\
Overall, despite a growing body of work on energy price forecasting and optimisation-based trading strategies, there remains a notable disconnect between forecast evaluation and algorithmic decision-making in practical BESS applications. Much of the existing literature focuses either on improving the accuracy of short-term electricity price forecasts \cite{improving_price_forecast_accuracy} or on developing sophisticated optimisation techniques for BESS scheduling \cite{using_LP_optimisation_in_BESS}, but rarely are these two strands of research integrated in a way that explicitly accounts for forecast reliability. This results in a gap between academic modelling and practical deployment. This thesis seeks to advance the literature by explicitly linking forecast performance to trading efficacy through a rigorous empirical assessment of AEMO forecast accuracy and its consequences for BESS arbitrage strategy outcomes. By using real operational forecasts from a system operator rather than back-tested or simulated data, this work captures realistic market behaviour and constraints. In doing so, it provides one of the first publicly documented, reproducible studies that not only evaluates AEMO forecasts across multiple temporal and regional dimensions, but also quantifies the financial returns of incorporating those forecasts into BESS control logic. This approach positions the thesis at the intersection of energy market forecasting, real-time optimisation, and applied energy economics, an area where transparent peer-reviewed research remains limited yet urgently needed for informed market participation and system design.

\section{Research Objectives and Contributions}

This research aimed to determine whether AEMO energy price forecasts could be systematically leveraged to develop a reliable and profitable BESS trading algorithm. To achieve this overarching aim, this study pursued a threefold set of objectives. First, it quantified the accuracy of AEMO energy price forecasts across all NEM regions to identify the most suitable locations for BESS deployment by using error metrics that captured variations by time of day, forecast horizon, and regional differences. This analysis provided critical insight into the patterns and limitations of forecast reliability and addressed a gap in the literature concerning the profitability of utilising publicly available market forecasts. Second, the study developed and validated a novel BESS trading algorithm designed to adapt its strategy based on observed forecast accuracy. Its performance was benchmarked against a baseline strategy that operated without forecast information which provided a comparative assessment of the added value forecasts bring to trading decisions. Finally, the research explored the potential for machine learning-driven enhancements to improve both forecast accuracy and BESS trading outcomes. Together, these objectives enabled a comprehensive assessment of how forecast quality impacts operational decisions and profitability in energy storage systems.
\\\\
The findings from this research have the potential to inform future developments in energy market trading models, and will offer insights that could facilitate more efficient BESS integration to improve overall market efficiency.

\section{Structure of the Thesis}

This thesis is structured into five chapters, progressively building an understanding of how AEMO price forecasts can optimise BESS trading strategies. From here onwards, Chapter \ref{ch: Methodology} details the methodology, including data sources, analytical techniques, and machine learning modelling employed to evaluate forecast accuracy and develop the trading algorithm. Chapter \ref{ch: Results} presents results by assessing AEMO forecast accuracy and comparing the financial performance of the developed algorithms, including machine learning-driven enhancements. Chapter \ref{ch: Discussion} discusses findings, and explores practical implications, limitations, and future research directions. Finally, Chapter \ref{ch: Conclusion} summarises the key contributions, highlights industry relevance, and offers recommendations, concluding with reflections on the research’s broader impact on BESS trading and market efficiency.



\begin{savequote}[80mm]
The greatest victory is that which requires no battle.
\qauthor{\textbf{---}Sun Tzu, \textit{The Art Of War}}
\end{savequote}

\chapter{Methodology}
\label{ch: Methodology}

\minitoc


\section{Research Approach and Framework Overview}
This research adopts a data-driven approach to evaluate the accuracy of AEMO energy price forecasts and develop a BESS trading algorithm that systematically incorporates forecast reliability. The study is conducted using Python to leverage its extensive ecosystem of numerical computing, data analysis, and machine learning libraries. The research follows a structured framework that integrates three key components: forecast accuracy assessment, algorithm development, and performance benchmarking.
\\\\
The first stage involves collecting and preprocessing historical AEMO price forecasts alongside actual market prices. By applying statistical analysis, the study quantifies forecast errors across different time horizons, regions, and market conditions. This analysis establishes the foundation for the trading strategy as it reveals patterns in forecast reliability that can be exploited for decision-making.
\\\\
The second stage focuses on algorithm development and begins with a baseline BESS trading model that operates without forecast information. This benchmark provides a control against which the accuracy-informed strategy can be evaluated. The novel trading algorithm incorporates the AEMO forecast by adjusting its decision-making based on observed price fluctuations. The model is designed to optimise arbitrage profitability by dynamically adjusting charging and discharging behaviour in response to predicted energy prices. Additionally, the study explores the use of machine learning to enhance AEMO forecasts and govern a more advanced trading strategy. The machine learning model is trained on historical market data and price trends to refine price predictions.
\\\\
The final stage involves performance benchmarking, where both the baseline and accuracy-informed algorithms are tested using historical market data. Key performance metrics such as financial returns and payback periods are computed and compared to hurdle rates, to assess the added value of incorporating AEMO forecasts. 
\\\\
By following this structured methodology, the research systematically evaluates the practical utility of AEMO forecasts in BESS trading and explores the potential of machine learning-driven enhancements. Python's extensive libraries, including Pandas \cite{Pandas} for data manipulation, Seaborn \cite{Seaborn} for visualisation, GurobiPy for optimisation \cite{GurobiPy}, and Scikit-Learn \cite{Scikit-learn} for machine learning, provide the computational foundation necessary to execute this study with precision and efficiency. All computations were executed on a machine equipped with 8 GB of RAM and a 7\textsuperscript{th} generation Intel Core i5 processor. This demonstrates that the proposed algorithms are lightweight and computationally efficient, with execution times comfortably within the NEM’s 5-minute settlement interval, perfectly adequate for deployment in real-world BESS operations.

\section{Data Sources and Preprocessing}
\label{sec:data_sources}

This study relies on historical market price data and AEMO forecast data to evaluate the accuracy of price predictions and their potential application in BESS trading strategies. The dataset includes actual market prices and forecasted prices across different time horizons to allow for a comparative analysis of forecast reliability and its impact on trading decisions.
\\\\
AEMO price data was retrieved programmatically using Python, with scripts automating the download, processing and merging of large historical datasets. The actual market prices were extracted from AEMO's dispatch price data \cite{DISPATCHPRICE}, while forecasted prices were obtained from AEMO's pre-dispatch data \cite{PREDISPATCHPRICE}, which contains price predictions made at different intervals ahead of real-time trading. Since the implementation of five-minute settlement in Australia's NEM on 1\textsuperscript{st} October 2021, generators are required to submit bids for each of the 288 five-minute trading intervals in a day, which replaced the previous 48 half-hour intervals \cite{NER200_3_8_4}. Initial bids must be submitted to AEMO by 12:30 PM on the day prior to dispatch. However, generators have the flexibility to adjust their bids through a process known as rebidding, which can be undertaken up until approximately three to four minutes before the relevant five-minute dispatch interval. This capability is particularly advantageous for BESS operators, as it allows them to update their bids in response to the latest market conditions and forecasts. Notably, while the market operates on five-minute settlement intervals, AEMO continues to publish pre-dispatch forecasts at 30-minute intervals. This discrepancy likely stems from legacy system constraints and has not been updated to align with the current five-minute settlement framework. The ability to rebid close to dispatch time enables BESS operators to leverage these forecasts effectively, and ensures that their bidding strategies are informed by the most recent data. This real-time responsiveness is crucial for the practical implementation of advanced bidding algorithms developed in this study, as it allows for dynamic adjustment to market conditions, thereby enhancing the economic viability and operational efficiency of BESS assets.
\\\\
All trading simulations were conducted using a standardised BESS configuration of 10 MW/20 MWh, equivalent to approximately five Tesla Megapacks, costing an estimated AU\$8M (US\$5M) \cite{MEGAPACK} however, excluding costs such as transformers, project development, balance of plant, and engineering, procurement and construction contractors. This choice was made to simulate reduced market power of a smaller BESS unit \cite{BESS_market_power} while still allowing for meaningful arbitrage opportunities. For example, in New South Wales, daily energy demand tends to be in the range of 6-10 GW \cite{AEMO_Dashboard}, hence, a 10 MW BESS will not have a significant impact on energy prices. In addition, although the Australian NEM operates on five-minute settlement periods \cite{AEMO_five_minute_settlement_periods}, the study adopts 30-minute intervals for both accuracy benchmarking and trading simulations. This adjustment aligns with AEMO’s 30-minute forecasting intervals \cite{PREDISPATCHPRICE}, reduces computational overhead, and mitigates file size constraints without significantly affecting market dynamics.
\\\\
Preprocessing involved several key steps: cleaning raw datasets to remove inconsistencies, aligning forecast timestamps with actual market prices, and caching data for efficient retrieval during simulations. To clean the datasets, frequency response market prices, cumulative energy prices, and regional override prices were removed, keeping only the regional reference price which was used to settle the market for each settlement period. Market price and forecast data were matched based on time, region, and forecast horizon to allow for a comprehensive evaluation of prediction accuracy and its impact on trading decisions. The final preprocessed dataset serves as the foundation for the subsequent analysis and ensures that BESS trading strategies can be assessed under realistic market conditions.

\section{Forecast Accuracy Analysis}

The accuracy of AEMO price forecasts plays a crucial role in determining their suitability for guiding BESS trading strategies. This study evaluates forecast reliability using statistical error metrics and examines the influence of key market factors, including time of day, forecast horizon, regional variations, and price volatility.
\\\\
To assess forecast reliability, historical AEMO price forecasts are compared against actual market prices. Two primary error metrics are used: absolute price error, which quantifies deviations in dollar terms, and percentage error, which normalises discrepancies relative to price magnitude. These metrics are evaluated across different market conditions to identify systematic patterns in forecast accuracy. Boxplots with 95\% confidence intervals of forecast error are generated to illustrate variations by hour of the day, day of the week, and month. These visualisations reveal whether accuracy follows predictable cycles and enables targeted adjustments in trading strategies.
\\\\
The forecast horizon is another critical factor influencing reliability. Errors tend to grow as the prediction window extends further into the future which reflects increased market uncertainty. This trend is captured through line plots that show mean forecast errors as a function of prediction lead time. By analysing these patterns, the study determines how far into the future AEMO forecasts remain useful for BESS decision-making.
\\\\
Regional variations in forecast accuracy are also examined, as different electricity markets experience distinct demand-supply dynamics, regulatory influences, and price-setting behaviours. Comparative analyses across NEM regions provides insights into which locations exhibit the most reliable forecasts, with the largest energy price fluctuations to guide regional selection for BESS deployment.
\\\\
Price volatility further influences trading decisions, as higher volatility presents greater arbitrage opportunities but also increases risk \cite{volatility_breeds_arbitrage}. This study quantifies within-day price volatility by computing the standard deviation of actual prices over 24-hour periods.
\\\\
The insights gained from this analysis inform the selection of the optimal NEM region for BESS deployment. By identifying market conditions where forecast reliability is highest and arbitrage opportunities are most profitable, the study ensures that the BESS trading algorithm is implemented in a manner that maximises financial returns.

\section{Development of the BESS Trading Algorithms}

\begin{figure}[!ht]
    \centering
    \includegraphics[width=0.8\textwidth]{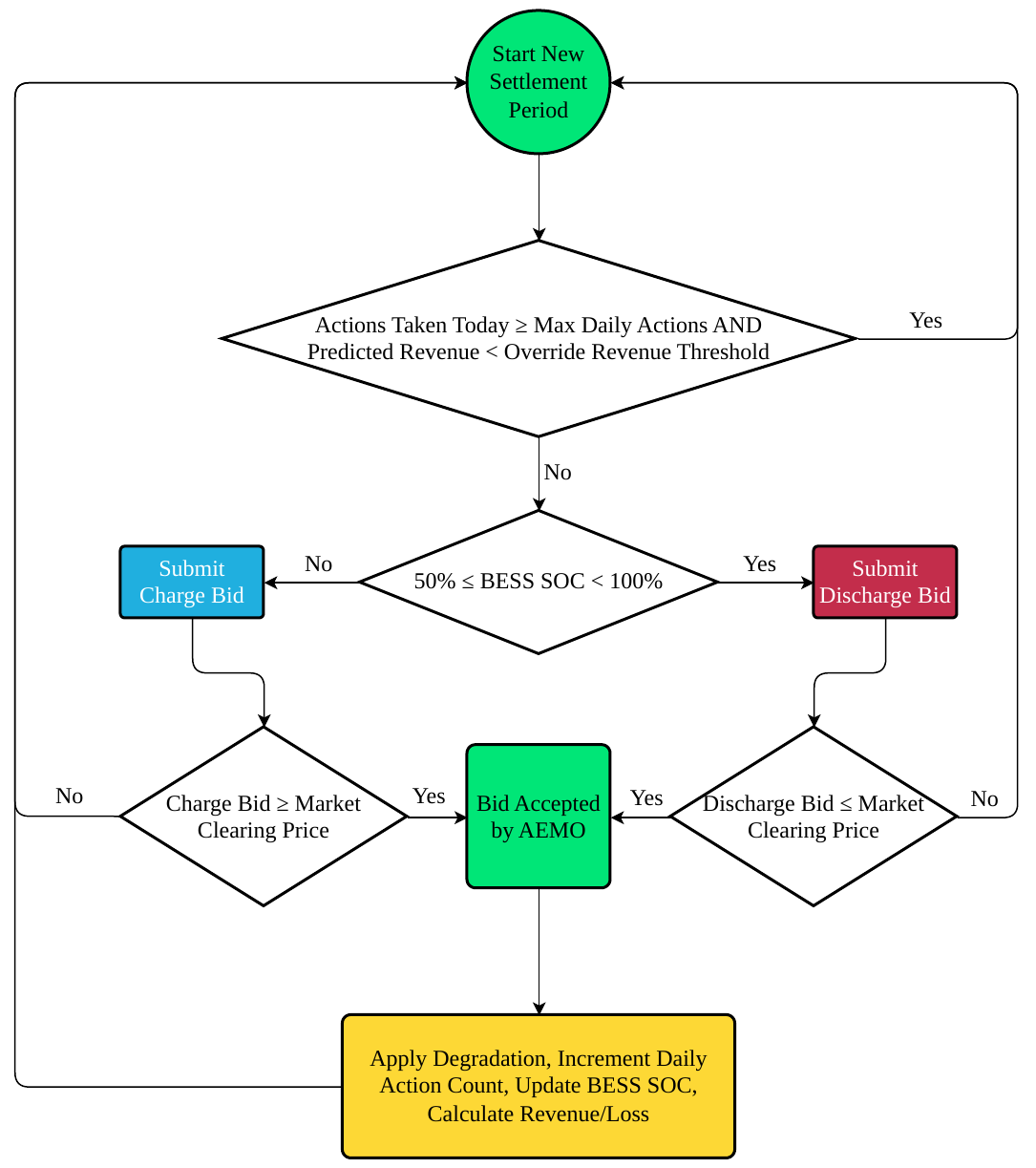}
    \caption{Threshold-Based Forecast-Unaware BESS Trading Strategy}
    \label{fig:simple}
\end{figure}

The development of the BESS trading algorithms was implemented in Python, focusing on two distinct approaches: a baseline algorithm that operates without forecast data and a forecast-aware algorithm that incorporates predictive insights to optimise trading decisions. Each algorithm's performance is assessed through extensive backtesting on every 30-minute settlement period of the 2024 historical energy market data.
\\\\
The baseline algorithm utilises a simple threshold-based decision-making process, whereby charging and discharging actions, defined in this study as any amount of deviation in the state of charge (\gls{SOC}), are determined solely by predefined price limits and the battery’s SOC. Specifically, when the SOC falls below 50\%, the algorithm places a bid to charge at a fixed price of \$50/MWh. Conversely, if the SOC exceeds 50\%, the system places a discharge bid at \$150/MWh. This binary logic eliminates the need for predictive analytics or external market forecasts and solely relies instead on instantaneous price observations and a static decision rule.
\\\\
To prevent excessive cycling and preserve operational realism, the algorithm imposes a daily action cap that limits maximum charge or discharge events per calendar day. Rudimentary battery degradation is also incorporated into the simulation as for each charge or discharge action, the total energy capacity is reduced by 0.005\%, modelling the gradual loss of storage potential over time due to wear and tear. For example, after 1,000 BESS actions in a year, corresponding to an average of approximately three actions per day, the maximum SOC is reduced to $(1 - 0.005\%)^{1000} \approx 95\%$. This relatively aggressive degradation rate is a realistic representation of what can be expected in the field, since 5\% loss in maximum SOC is typically seen after 500 cycles \cite{degradation_vs_cycles}, where a cycle is a full charge and discharge of the BESS. Annual BESS cycles vary wildly and can range from 147 \cite{cycles1} to 254 \cite{cycles2} cycles per year, hence, this simple degradation model provides a lower-bound estimate of the potential trading revenue. It also compensates for the omission of round-trip efficiency losses, which are not considered in this study. A more comprehensive analysis of battery degradation, incorporating temperature effects, depth-of-discharge cycling, and efficiency losses, is recommended for future research, as complex degradation modelling was not within the scope of this research. The inclusion of a simple degradation model ensures that the long-term performance of the battery system is reflected in the trading outcomes.
\\\\
The forecast-unaware baseline algorithm depicted in \autoref{fig:simple} and further detailed in \autoref{alg:simple}, was applied to historical NEM data, which executed trades depending on the actual market price and the bid thresholds, to accurately simulate real accepted or rejected bids by the market operator. The historical price for each settlement period was assumed to be the clearing price of that period, given the negligible market impact from a 10 MW BESS. A charge action was triggered only when the market clearing price was below or equal to the charge bid, ensuring economically viable energy purchases. Likewise, a discharge action was executed only if the market clearing price met or exceeded the discharge bid, thereby guaranteeing positive revenue generation. This simple mechanism, although lacking adaptive intelligence or forecast awareness, serves as a foundational benchmark for evaluating the added value of more complex, forecast-driven optimisation strategies introduced later in this study.

In contrast, the forecast-aware algorithm integrates predicted electricity prices to inform trading decisions. This model considers a 24-hour lookahead period, during which it evaluates the predicted price trajectory to determine the most profitable moments to charge or discharge. Instead of relying on fixed bid thresholds, the algorithm dynamically adjusts its buy and sell decisions based on forecasted price extrema, seeking to charge when future prices are expected to rise and discharge when prices are projected to peak. At each settlement interval, it extracts all available forecasts made prior to the current time, with horizons extending up to the lookahead window. A weighted average of predicted prices is computed for each future settlement period, where more recent forecasts are given greater weight. These values are then used in a linear programming (\gls{LP}) optimisation \cite{LP_optimisation_in_BESS} that maximises total revenue across the lookahead horizon, subject to constraints on battery capacity, power limits, and daily action limits outlined in Section \ref{sec:milpoptimisation}. The solution to this optimisation problem yields an action, \textit{Charge}, \textit{Discharge}, or \textit{Hold}, to be taken at the current time step.
\\\\
Additionally, a revenue-based override mechanism is implemented, enabling the algorithm to execute further trades beyond the daily action limit if the expected revenue exceeds a predefined threshold (e.g., \$1,000). This allows the model to exploit highly profitable opportunities that would otherwise be missed due to operational constraints. An alternative approach to achieve a similar result would be to incorporate battery degradation as a penalty term in the objective function, however, this was not explored since firstly, it would increase the complexity of the objective function making it slower to compute for every time step and secondly, more advanced degradation modelling was outside the scope of this project.

\section{Optimising BESS Trading Decisions via Mixed-Integer Linear Programming}
\label{sec:milpoptimisation}

To systematically optimise trading decisions for the forecast-aware BESS algorithms, this study formulates the problem as a Mixed-Integer Linear Program (\gls{MILP}). An MILP is an optimisation framework that includes both continuous and integer decision variables, governed by a linear objective function and a set of linear constraints. In this formulation, charging and discharging power levels are modelled as continuous variables, while binary variables represent mutually exclusive operational decisions, being charge, discharge, or hold at each time step. The combination of variable types and the linear structure of the objective and constraints makes this problem a MILP. This framework provides a robust and flexible foundation, enabling the integration of both AEMO price forecasts and machine learning-enhanced forecasts. The entire optimiser is developed using Python, specifically employing \texttt{pandas} and \texttt{NumPy} for data manipulation and numerical operations. At first, the \texttt{PuLP} library for linear optimisation was utilised for model setup and testing. However, it became apparent that \texttt{PuLP} would be inadequate to process large datasets, hence, the \texttt{GurobiPy} package was deployed under an academic licence to greatly hasten the optimisation of trading decisions. The optimiser runs in a time-stepped loop across the entire dataset, dynamically constructing and solving a new MILP problem for each settlement period using only the information that would have been available at that point \cite{using_LP_optimisation_in_BESS}. This simulates real-world conditions of imperfect foresight and ensures that the backtesting results reflect practical feasibility. The following outlines the methodology behind this optimiser, detailing how forecast data is processed, transformed, and subsequently fed into a constrained MILP model.
\\\\
At the core of this method lies a time-stepping framework which iterates over each 30-minute settlement period in the dataset. For each period, the optimiser considers all AEMO price forecasts that were available up to that moment and for a fixed lookahead window, of 24 hours into the future. These forecasts are weighted inversely with respect to their time-ahead values, assigning greater weight to forecasts made closer to the target settlement period. This weighting reflects the empirical reality that forecast reliability typically increases with proximity to the dispatch time \cite{inverse_hour_ML_price_weighting}, and follows from the observations made from the accuracy assessment based on how far ahead in time the predictions were made.
\\\\
Let $t$ denote the current settlement period, and let $\mathcal{T} = \{t_1, t_2, \ldots, t_n\}$ be the set of future periods within the lookahead window. For each $t_i \in \mathcal{T}$, a weighted average forecast price is computed, $P_i$, where more recent forecasts are given greater weight, using all available 30-minute forecasts made prior to or at time $t$ for $t_i$. These prices form the economic basis for the MILP model which selects the optimal charging or discharging action for each period in $\mathcal{T}$.
\\\\
The goal of the MILP is to maximise profit, defined as the difference between revenue earned from discharging and costs incurred from charging over the lookahead window. Let $P_i$ represent the predicted price (in \$/MWh) for settlement period $i$, $c_i$ the energy charged (MWh), and $d_i$ the energy discharged (MWh). Let $h$ denote the settlement period duration in hours (e.g., $h = \frac{1}{2}$ for 30-minute intervals). The objective function is therefore:

\begin{equation}
\max \sum_{i=1}^n \left( P_i \cdot d_i - P_i \cdot c_i \right) \cdot h
\end{equation}

Subject to the following constraints:

\textbf{Energy balance and state-of-charge (SOC):} For each period, the battery SOC is updated as:
\begin{equation}
SOC_i = SOC_{i-1} + c_i \cdot h - d_i \cdot h
\end{equation}
\begin{equation}
0 \leq SOC_i \leq C_{\text{max}}
\end{equation}
where $C_{\text{max}}$ is the maximum energy capacity of the battery.

\textbf{Power constraints:} Charging and discharging power are bounded:
\begin{equation}
0 \leq c_i \leq P_{\text{max}}, \quad 0 \leq d_i \leq P_{\text{max}}
\end{equation}
where $P_{\text{max}}$ is the rated power (in MW) of the battery.

\textbf{Charge/discharge thresholds:} Charging is only permitted when $P_i \leq \theta_{\text{charge}}$ and discharging when $P_i \geq \theta_{\text{discharge}}$:
\begin{equation}
c_i \leq P_{\text{max}} \cdot [P_i \leq \theta_{\text{charge}}]
\end{equation}
\begin{equation}
d_i \leq P_{\text{max}} \cdot [P_i \geq \theta_{\text{discharge}}]
\end{equation}
where $\theta_{\text{charge}}$ and $\theta_{\text{discharge}}$ are charge and discharge price thresholds, respectively, set from the sensitivity analysis heatmap discussed in Section \ref{sec:simple}. The inclusion of price thresholds in the optimiser is essential because ultimately, the BESS must submit a discrete market bid to AEMO. While the MILP optimiser determines whether to charge, discharge, or hold based on forecasted prices, it does not directly generate a price bid. To bridge this gap, the thresholds derived from the sensitivity analysis are used to guide these discrete operational decisions within the MILP framework. Without these thresholds, if the optimiser were instead tasked with determining an optimal bidding price, it would likely overestimate the market clearing price, consistent with the overprediction tendencies observed in Section \ref{sec:accuracy}. This would result in a higher frequency of rejected discharge bids, thereby reducing overall revenue.

\textbf{Mutually exclusive actions:} Only one action, charge, discharge, or hold, is allowed per time step:
\begin{equation}
c_i + d_i \leq P_{\text{max}} \cdot (1 - h_i)
\end{equation}
where $h_i$ is a binary variable indicating a hold action.

\textbf{Daily action limit:} The number of active decisions (charge or discharge) per calendar day is capped:
\begin{equation}
\sum_{i \in D} a_i \leq A_{\text{max}} - A_{\text{used}, D}
\end{equation}
where $a_i$ is a binary variable representing an active decision at period $i$ on date $D$, $A_{\text{max}}$ is the permitted maximum number of actions per day, and $A_{\text{used}, D}$ is the number already executed.
\\\\
The optimiser solves this MILP for each settlement time by using the weighted price predictions over the defined horizon. At each iteration, the MILP returns the optimal decisions of whether to charge, discharge, or hold for each settlement period in the next 24 hours. These decisions are stored and subsequently used to guide real-time trading behaviour. If a forecasted revenue does not exceed a predefined override threshold and the daily action limit has been reached, the algorithm withholds any action to avoid suboptimal engagement and to enhance battery lifespan.
\\\\
Once the optimal decisions have been identified for each lookahead horizon, only the immediate next step is executed. This ensures a rolling-horizon approach where the battery state is updated at every settlement period based on real market prices, while future actions are continuously re-evaluated. The battery’s state-of-charge is updated by accounting for the energy charged or discharged during the interval. Charging incurs a cost based on the actual market price, while discharging earns revenue. These monetary transactions are accumulated to determine the total arbitrage profitability over the analysis period.
\\\\
This linear programming-based optimiser serves as the core decision engine for BESS arbitrage trading in this study. It balances predictive insight with operational constraints and utilises weighted energy price forecasts to dynamically determine profitable charge and discharge schedules. By solving a constrained optimisation problem at each timestep and enforcing practical limits on actions and state-of-charge, the model emulates a realistic BESS trading system. The results from this optimiser are later benchmarked against a baseline strategy with no access to forecast data which allows for quantification of the value added by forecast-informed optimisation.

\section{Machine Learning Integration for Price Forecasting}

The integration of machine learning into the price forecasting pipeline is a critical component of this methodology and serves as the foundation for informed bidding and operation decisions of the BESS. A common strategy is to combine machine learning to predict energy prices with MILP optimisation to determine the most profitable trading decision \cite{ML_plus_LP}. While the downstream MILP optimiser ultimately governs the precise schedule of charging and discharging actions, the success of such decisions is predicated on the quality and granularity of short-term electricity price predictions. This section outlines the machine learning pipeline responsible for generating enhanced forecasts using a supervised learning approach, specifically leveraging a Random Forest Regressor model \cite{random_forest_to_forecast} trained once on the 2023 historical price dataset and tested on 2024 data.

\vspace{20pt}

\begin{figure}[!ht]
    \centering
    \includegraphics[width=0.8\textwidth]{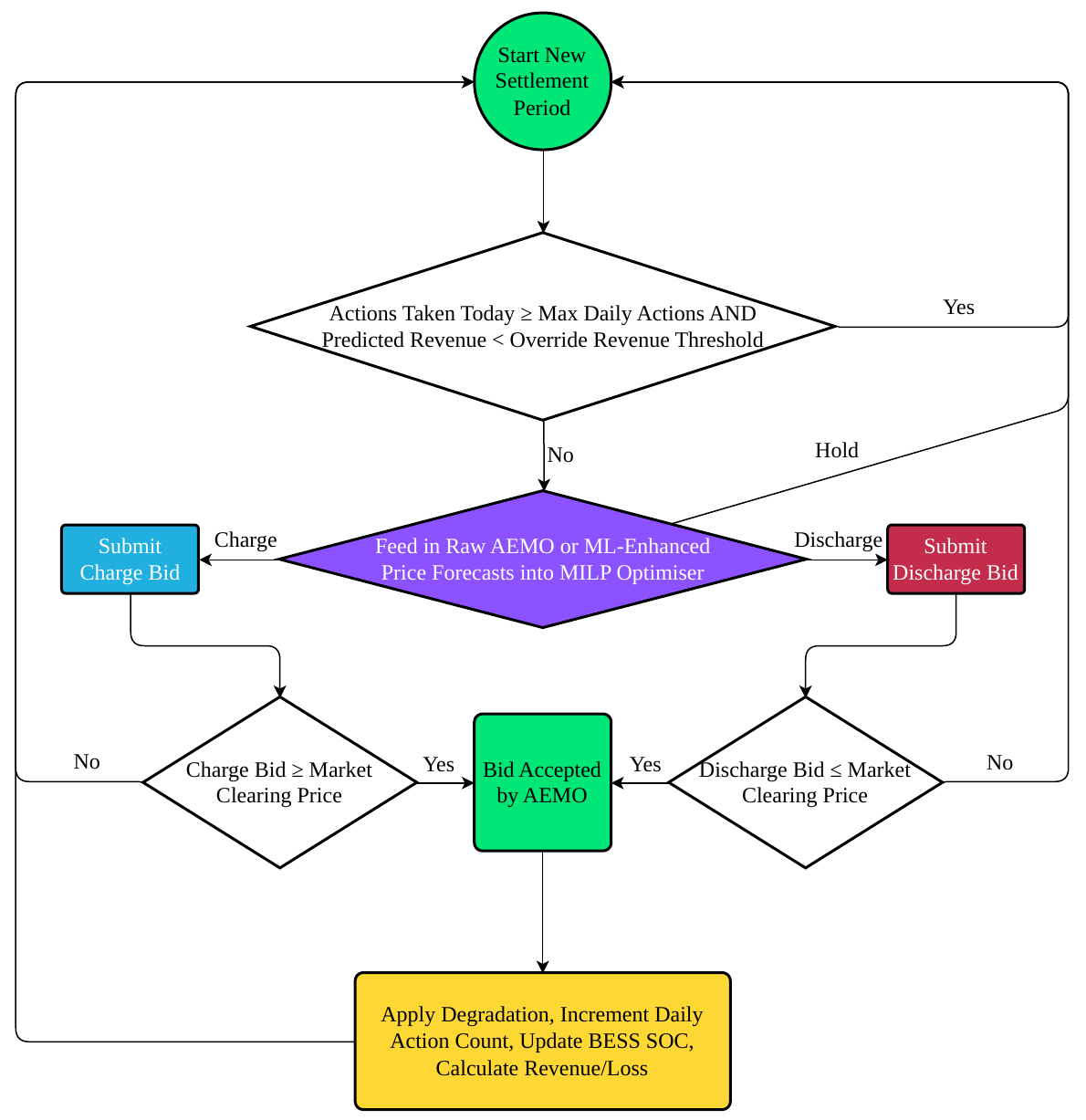}
    \caption{MILP-Optimised Algorithm with ML-Enhanced Price Forecasts}
    \label{fig:ml}
\end{figure}

To begin with, all time-related data was transformed into a consistent \texttt{Pandas} datetime format, ensuring alignment across features that involved price settlement times and the time when a prediction was made. From this, a derived field \texttt{When Prediction Was Made} was calculated by subtracting the time horizon of each forecast from its respective settlement time. This operation allowed the model to account for the temporal distance between the prediction and its target period, a key consideration for weighting forecast confidence, and ensured that the prediction model only had access to data it would have up until the latest time step.
\\\\
In the feature engineering stage, temporal features were constructed to capture cyclical behavioural patterns inherent in electricity demand and pricing \cite{sin_cos_transformations}. Using the hour and minute components of the \texttt{Settlement Date \& Time}, a \texttt{TimeOfDay} variable was derived and decomposed using sine and cosine transformations to preserve the cyclical nature of the 24-hour clock. Similarly, the day of the week was extracted from the same timestamp and converted into \texttt{DayOfWeek\_sin} and \texttt{DayOfWeek\_cos} to retain the periodic structure of weekly trends. These cyclical transformations prevent discontinuities at boundaries such as midnight or the start of the week to ensure smoother modelling of time-series behaviour. The final feature set used for modelling included these engineered cyclic variables alongside the \texttt{AEMO Predicted Price (\$/MWh)} and \texttt{Hours Ahead of Prediction}, all of which are logically relevant predictors of short-term price movement.
\\\\
The model is trained on six features: AEMO predicted price, hours ahead of prediction, and the four cyclical features capturing the sine and cosine of the time of day and day of week. The target variable was the actual price observed in the market for each settlement period. After fitting the model, its performance was evaluated on the training data using root mean squared error as a diagnostic measure, and the ML-enhanced forecast accuracy was plotted to compare against the AEMO forecast.
\\\\
The system is designed to operate in a rolling prediction environment, where model forecasting is done dynamically across the test period. The decision to predict prices every settlement period is based on the assumption that recent data offers more relevant insights into price behaviour, particularly in markets like Australia's NEM where price dynamics can shift rapidly due to weather events, supply-demand fluctuations, or policy interventions. For each unique settlement period in the test set, the system constructs a forward-looking prediction set comprising all forecasts made at or before the current time, and that correspond to a settlement period within the specified lookahead window, as dependent on the AEMO price forecasts. These candidate forecasts are then passed through the trained model to obtain ML–enhanced predicted prices. To ensure that nearer-term predictions are treated as more reliable than those made further ahead, the same weighting scheme is applied as in Section \ref{sec:milpoptimisation}. 
\\\\
Although the MILP component governs action feasibility and scheduling, the quality of the underlying forecasts has significant downstream impact on operational and financial outcomes. Forecasts that under- or overestimate prices can result in suboptimal arbitrage, either by missing lucrative discharge opportunities or by charging at inflated prices. For this reason, a critical part of the integration involves dynamic tracking of decision quality. Following each optimiser execution, decisions for the immediate settlement period are recorded and implemented if they pass revenue thresholds and action constraints.
\\\\
Further, the final bid decision is conditional not only on the optimiser’s prescribed action but also on real-world market conditions such as the actual price and the battery’s SOC. In instances where the optimiser recommends charging or discharging and the predicted revenue justifies it, the aforementioned override mechanism permits the action even if the daily action limit has been exceeded. This is particularly important in high-volatility scenarios, where large price spikes could significantly enhance revenue if captured promptly.
\\\\
Overall, this methodology is outlined in \autoref{fig:ml} and is further detailed in \autoref{alg:ml}. It integrates machine learning with rolling forecasts and temporal feature engineering to produce high-quality, short-term electricity price forecasts that guide the operation of a grid-connected BESS. The system leverages recent historical data and applies domain-informed cyclical transformations to better model the underlying temporal structure of the market. Through prediction aggregation via inverse-hour weighting, the approach ensures that the most relevant and timely information is used for operational decision-making. This architecture is well-suited to electricity markets where predictability varies substantially over time and where flexible and responsive strategies can yield substantial economic returns. The effectiveness of this machine learning component is inherently linked to the performance of the optimisation stage which makes its role foundational to the broader control strategy.

\section{Performance Benchmarking and Evaluation}
To evaluate the effectiveness of the BESS trading strategies, several key performance metrics were analysed, including revenue per year, cumulative lifetime revenue, and payback period, alongside a discounted cash flow analysis.
\\\\
For the forecast-unaware baseline algorithm, a sensitivity analysis was conducted to explore the impact of different buy and sell thresholds on trading outcomes. A heatmap was generated to visualise how varying charge and discharge price thresholds affect total revenue. By systematically testing different bid price combinations within a specified range (e.g., buy thresholds from -\$50/MWh to \$50/MWh and sell thresholds from \$50/MWh to \$150/MWh), an optimal threshold combination was identified that maximised trading profits. The results of this threshold optimisation were presented in a heatmap, where each grid cell represents the total revenue achieved for a given buy/sell threshold pair. The highest revenue configuration could then be identified to demonstrate the most profitable static trading strategy under historical market conditions.
\\\\
For the forecast-aware algorithm, performance was measured by comparing its revenue generation against the baseline strategy. A cumulative revenue plot was used to track the financial progression of each approach over time. The analysis also considered operational efficiency metrics such as the average number of trades per day. The inclusion of predictive insights allowed for improved trade timing, resulting in more optimal charge and discharge cycles.
\\\\
Financial return analysis of the BESS under each trading strategy was explored by comparing annual income and expenses with the initial capital investment cost to find key project finance metrics. The results provided insight into how forecast-driven trading affects the financial viability of battery storage systems. In addition to calculating income and expenses, key discounted cash flow metrics such as Net Present Value (\gls{NPV}) and Internal Rate of Return (\gls{IRR}) were utilised to assess long-term profitability. NPV accounts for the time value of money and reflects the present-day worth of future cash flows after considering discounting, whilst IRR represents the discount rate at which the project breaks even in present value terms. These indicators are particularly relevant for industry as they allow a more rigorous and investment-oriented comparison of trading strategies by quantifying the extent to which each algorithm can generate value over the battery’s operating lifetime relative to the upfront expenditure.
\begin{savequote}[\textwidth/2]
Every tree that bringeth not forth good fruit is hewn down, and cast into the fire.
\\
Wherefore by their fruits ye shall know them.
\qauthor{\textbf{---}Matthew 7:19\textbf{--}20}
\end{savequote}

\chapter{Results}
\label{ch: Results}

\minitoc


\section{Accuracy Assessment of AEMO Energy Price Forecasts}
\label{sec:accuracy}

\begin{figure}[!ht]
    \centering
    \begin{subfigure}[c]{0.495\textwidth}
        \includegraphics[width=\textwidth]{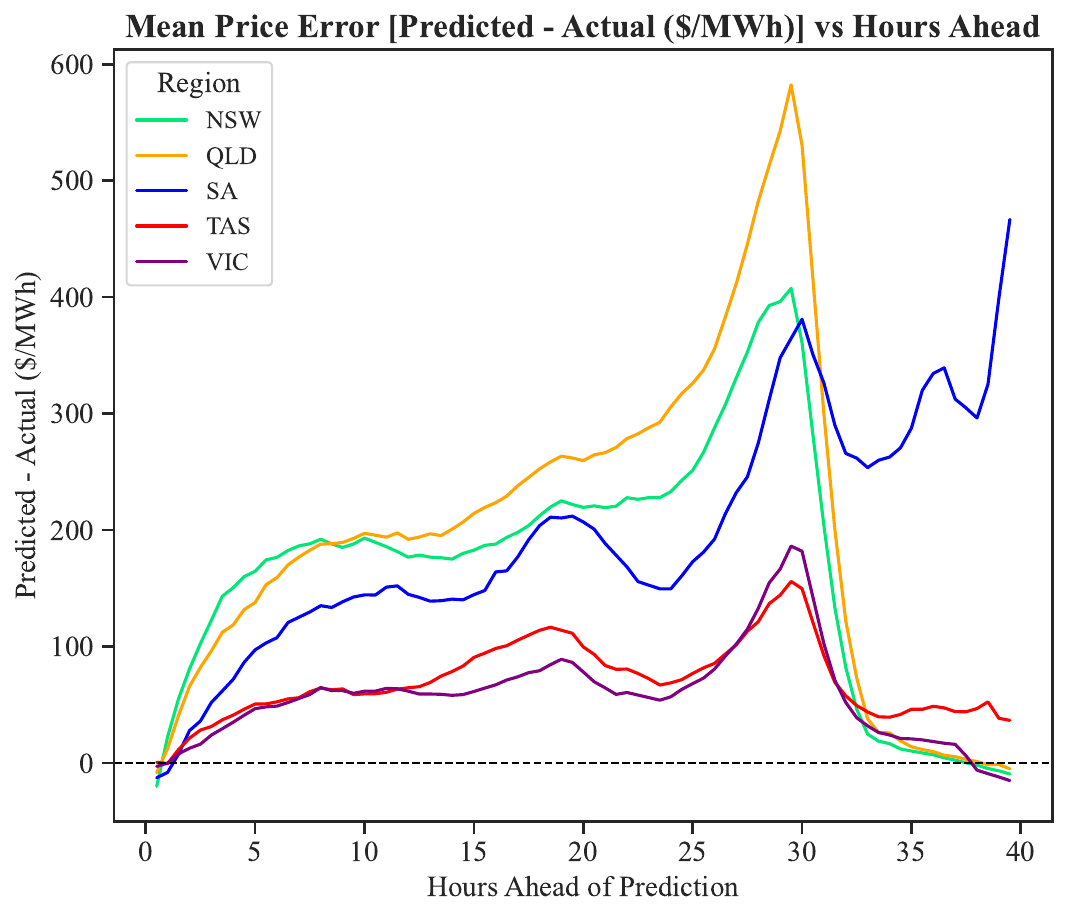}
        \caption{Mean Price Error}
        \label{fig:AEMO_mean_error}
    \end{subfigure}
    \hfill
    \begin{subfigure}[c]{0.495\textwidth}
        \includegraphics[width=\textwidth]{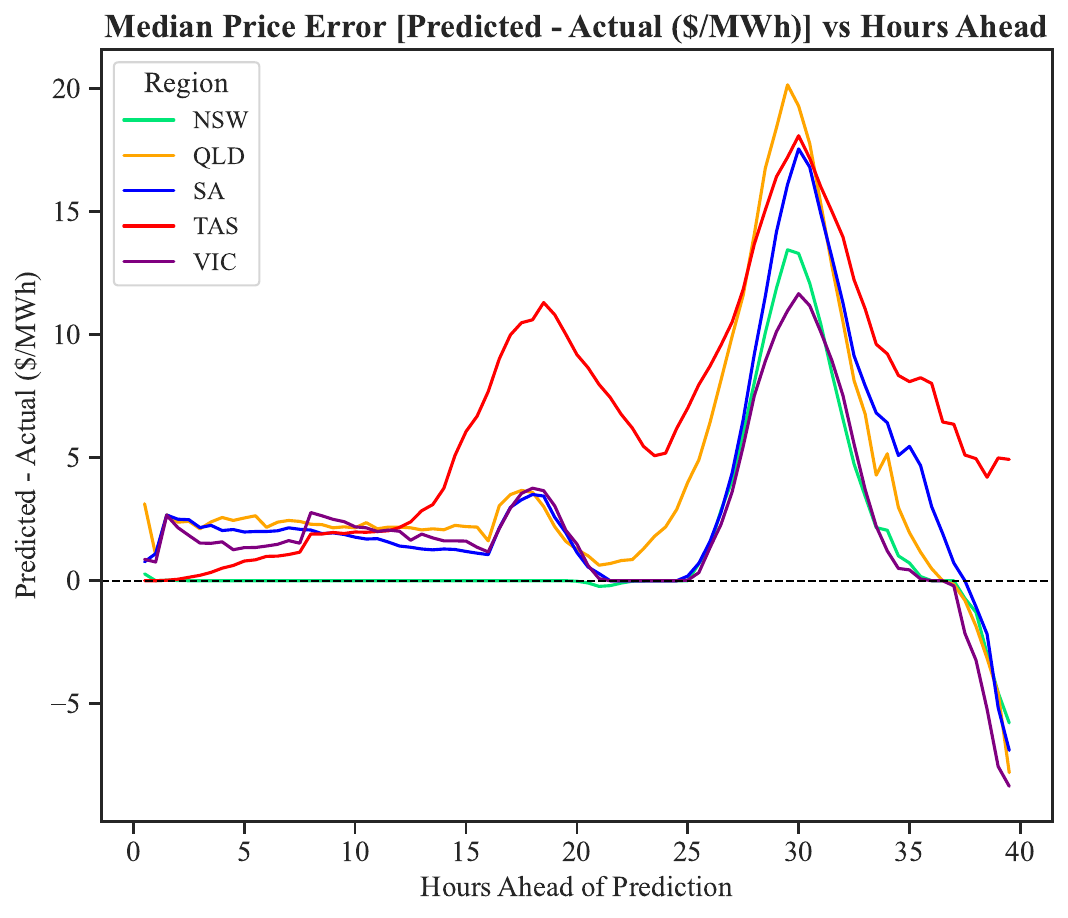}
        \caption{Median Price Error}
        \label{fig:AEMO_median_error}
    \end{subfigure}
    
    \caption{AEMO Forecast Regional Accuracy Ahead of Prediction}
    \label{fig:Regional_Accuracy}
\end{figure}

To evaluate the suitability of AEMO’s price forecasts for informing BESS trading strategies, this section provides a rigorous analysis of forecast accuracy across temporal horizons and regional contexts. Forecast errors are assessed across the five mainland NEM regions which frequently exhibit volatility \cite{NEM_volatility_spillovers}: New South Wales (\gls{NSW}), Queensland (\gls{QLD}), South Australia (\gls{SA}), Victoria (\gls{VIC}), and Tasmania (\gls{TAS}). Based on 2024 data, the analysis reveals both the magnitude and distribution of prediction errors over time, to identify difficult-to-predict regions and inform the design of the trading algorithms developed later in this thesis.
\\
The plot of mean error versus forecast horizon in \autoref{fig:AEMO_mean_error} reveals an expected and intuitive trend, as the forecast horizon increases, the accuracy of AEMO’s predictions deteriorates. Very short-term forecasts close to 1 hour tend to be relatively accurate, while errors tend to grow with each additional hour of forecast lead time. Interestingly, however, there is a surprising and sharp reduction in mean error beyond the 30-hour mark. This behaviour is observed across all regions with the exception of SA. A plausible explanation for this anomaly is that AEMO’s forecasting model may undergo a methodological shift beyond 30 hours, potentially relying less on dynamic variables such as weather data and instead reverting to static or historical pricing patterns. The discontinuity at this point supports the hypothesis that forecasts beyond 30 hours are generated using a different underlying mechanism, one possibly less sensitive to the unpredictable drivers of short-term price fluctuations. However, SA is relatively isolated within the NEM, with fewer and more constrained interconnectors to other regions. This limited ability to buffer local supply-demand imbalances through interregional trade can exacerbate price volatility and make the market more sensitive to forecast deviations. Region-specific analysis shows QLD has the highest absolute forecast error across most horizons, likely due to its larger geographical area, diverse generator mix, or transmission constraints.

\begin{figure}[!ht]
    \centering
    \includegraphics[width=1\textwidth]{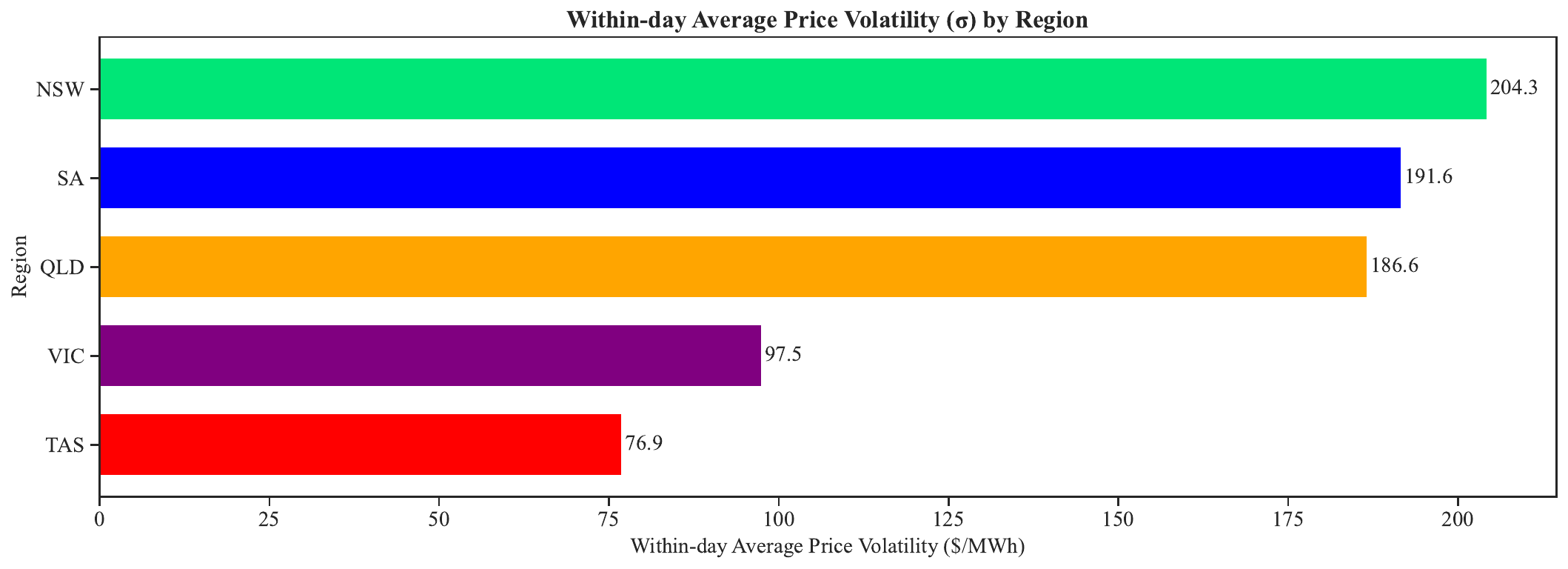}
    \caption{AEMO Forecast Regional Volatility}
    \label{fig:Regional_Volatility}
\end{figure}

In contrast, the median price error analysis in \autoref{fig:AEMO_median_error} provides a more robust view of forecast performance by mitigating the influence of extreme price spikes that distort mean-based metrics. Whereas the mean error plot showed values reaching up to \$600/MWh, particularly in QLD, where price volatility and occasional outliers significantly skewed the results, the median error remains tightly bounded between -\$10/MWh and \$20/MWh across all regions. This narrower range suggests that, for the majority of forecast instances, AEMO’s predictions are reasonably centred around actual market prices. Notably, NSW exhibits exceptional forecast performance for forecasts made between half an hour and 25 hours ahead, the median error remains consistently close to zero. This indicates a remarkably unbiased forecast for NSW over a substantial forecast horizon and suggests strong model calibration in this region.
\\\\
A consistent drop in median error is observed across all regions beyond the 30-hour mark which aligns with the earlier hypothesis that AEMO’s forecasting methodology shifts at this point, possibly moving from dynamic, real-time inputs to static or historical modelling techniques. This shift introduces a systematic downward bias, reflected in the negative median errors reaching -\$10/MWh at longer horizons. Interestingly, TAS diverges from this trend by maintaining a slightly positive median error up to 40 hours ahead. This may be due to the unique supply-demand structure of TAS, which is heavily reliant on hydroelectric generation and is weakly interconnected with the rest of the NEM which reduces exposure to shared price shocks and makes historical patterns more persistent. Another subtle yet intriguing feature is the bump in median error observed in all regions between 15 and 20 hours ahead. This temporary error increase may stem from the transition between intra-day and day-ahead strategies in AEMO’s forecasting and the interplay between ramping constraints and peak demand prediction. TAS experiences the largest deviation during this window, likely due to its limited market liquidity and smaller data volume, whilst NSW remarkably maintains a zero median error throughout which reinforces the exceptional stability of forecasts in that region.
\\
Overall, the median error analysis supports the conclusion that NSW offers the most forecastable market conditions, with minimal systemic bias and reasonable mean price error. This positions NSW as a strong candidate for developing and deploying the accuracy-informed BESS trading algorithm.
\\\\
These observations transition naturally into an analysis of volatility which serves as a proxy for the inherent difficulty of forecasting prices in each region, and therefore represents opportunity for larger energy price spreads to be captured by arbitrage algorithms. Within-day average price volatility in \autoref{fig:Regional_Volatility} was calculated using the standard deviation of intraday prices and highlights the erratic nature of the energy market, where sharp fluctuations occur on a daily basis. The price volatility graph further quantifies this uncertainty, showing that NSW, QLD, and SA are significantly more volatile than VIC and TAS. NSW, in particular, not only exhibits the highest within-day average price volatility but also shows relatively moderate forecast error, compared to SA and QLD, which makes it a compelling target region for the development and deployment of BESS trading strategies. Its combination of volatility and forecast reliability ensures there is enough arbitrage opportunity while still allowing for model-informed decision-making. On this basis, NSW is selected as the focus region for the development of trading algorithms in the remainder of this thesis.

\begin{figure}[!ht]
    \centering
    \includegraphics[width=1\textwidth]{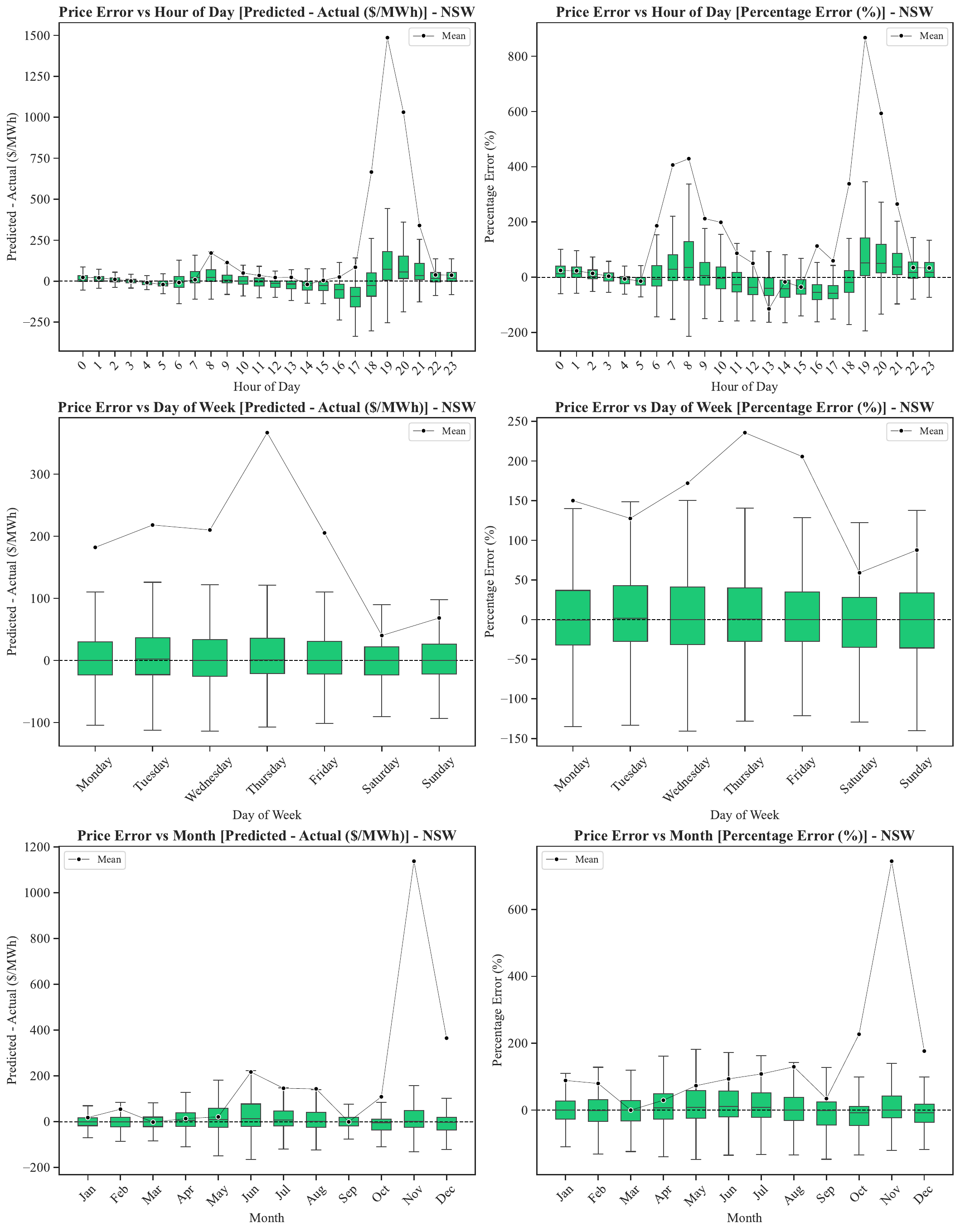}
    \caption{AEMO NSW Forecast Accuracy}
    \label{fig:NSW_accuracy}
\end{figure}

Having identified NSW as the target region, the analysis now narrows its focus to investigate forecast accuracy within NSW in more detail. \autoref{fig:NSW_accuracy} provides two columns, error in absolute terms (\$/MWh) and error as a percentage, and three rows corresponding to temporal groupings: hour of day, day of week, and month of year. This multidimensional breakdown provides a nuanced understanding of when the AEMO forecasts are most and least reliable.
\\

The first row, showing forecast error against hour of day, reveals a clear temporal structure in the forecasting performance. Errors are lowest during the late night and early morning hours, when demand is typically lowest and market conditions are more stable. However, forecast errors increase during the morning peak around 8:00 am and sharply in the evening around 7:00 pm, both periods of heightened and volatile demand. During these hours, AEMO frequently overestimates prices, likely due to the inherent uncertainty in predicting demand surges and generator availability in real time. Furthermore, during late afternoons, the rapid decline in solar output combined with the increased demand from energy customers necessitates a swift response from flexible generators such as gas peaking plants. However, due to limited participation of truly responsive capacity and the high marginal cost of fast-ramping units, the system operates with a constrained supply margin, often resulting in increased prices as the market adjusts. Interestingly, there is a tendency for AEMO to slightly underestimate prices around midday, a period that often coincides with high solar output and thus more depressed wholesale prices, indicating a possible overestimation of the impact of solar generation on net demand.
\\\\
The second row explores variation by day of the week. AEMO forecasts generally overestimate prices, with reduced errors on weekends. This supports the expectation that weekends have more stable, lower demand profiles that are easier to predict. Low weekend error suggests reduced demand variability improves forecast accuracy. Notably, Thursdays exhibit the highest forecast errors, which coincides with the day when 45\% of people engaged in remote working return to the office in NSW \cite{usyd_tops}.
\\
The third and final row in this figure charts forecast error across the months of the year. One might expect the error to be highest during summer (December to February), when peak loads place stress on the grid, but the data tells a more nuanced story. Forecast errors are relatively low from January through March, even as temperatures rise. It is in the spring-to-summer transition, specifically October to December, that forecast error climbs sharply. This anomaly may be explained by the unpredictability of weather during these months in NSW \cite{NSW_climate_change_unpredictability}, with sudden shifts between heatwaves and cool changes \cite{cool_change} introducing volatility that AEMO’s model struggles to capture. The consistent overestimation during this period suggests that the forecasting system may be anticipating higher prices in response to expected heat-driven demand, but is unable to fully account for the moderating effects of intermittent renewable generation or cooler-than-expected days.
\\\\
In summary, this analysis provides strong evidence that while AEMO’s forecasts exhibit reasonable accuracy at shorter horizons, they are subject to distinct biases and limitations that vary across time and region. In NSW, forecast accuracy is highest during low-demand periods and deteriorates during peak hours and seasonally volatile months. AEMO’s consistent tendency to overestimate prices, particularly in transitional periods, offers a predictable bias that can be potentially exploited by an adaptive trading strategy. Although the mean error remains consistently high, the median errors in both the day of week and month-based plots are tightly centred around zero, suggesting that a small number of large outliers likely driven by rare but extreme market events are skewing the mean upwards. These insights will directly inform the design of the machine learning models in later chapters, which will use hour of day and day of week as key predictive features to refine and enhance forecast-informed trading decisions.

\vspace{100pt}

\section{BESS Trading Performance Comparison}
\label{sec:trading_performance}

\begin{figure}[!ht]
    \centering
    \begin{subfigure}[b]{1\textwidth}
        \centering
        \includegraphics[width=\textwidth]{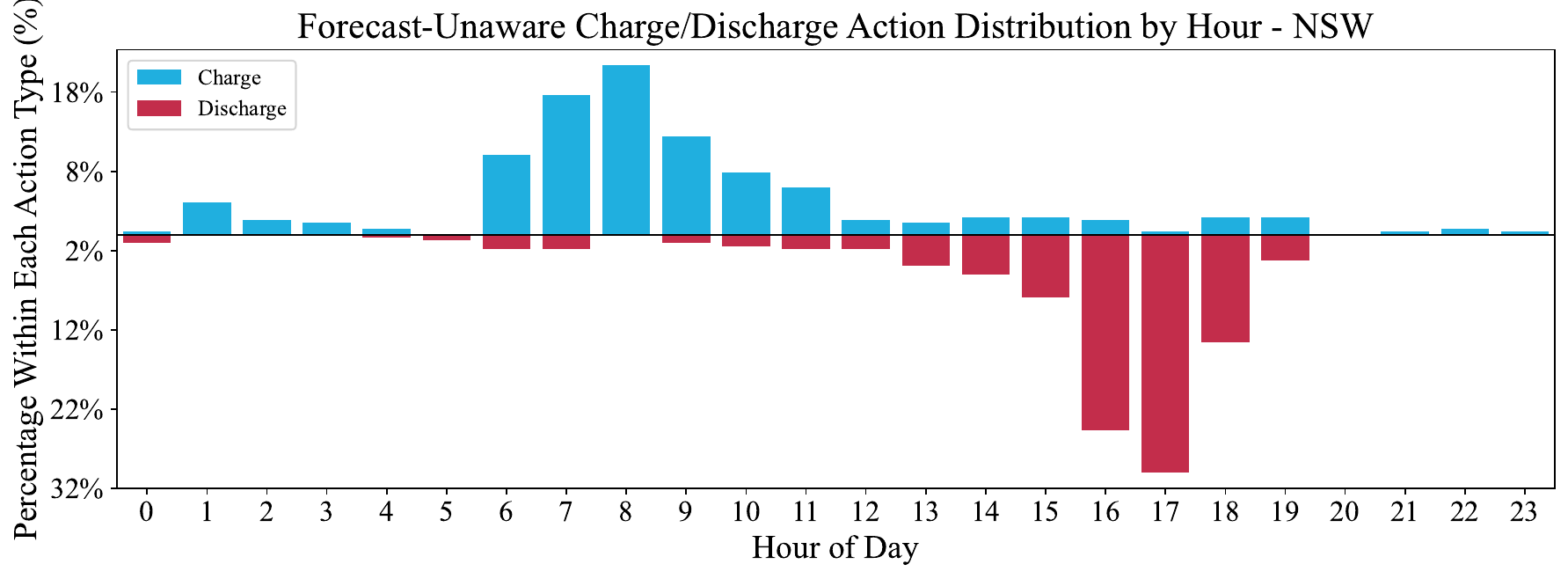}
        \caption{Forecast-Unaware Algorithm}
        \label{fig:simple_histogram}
    \end{subfigure}
    
    \begin{subfigure}[b]{1\textwidth}
        \centering
        \includegraphics[width=\textwidth]{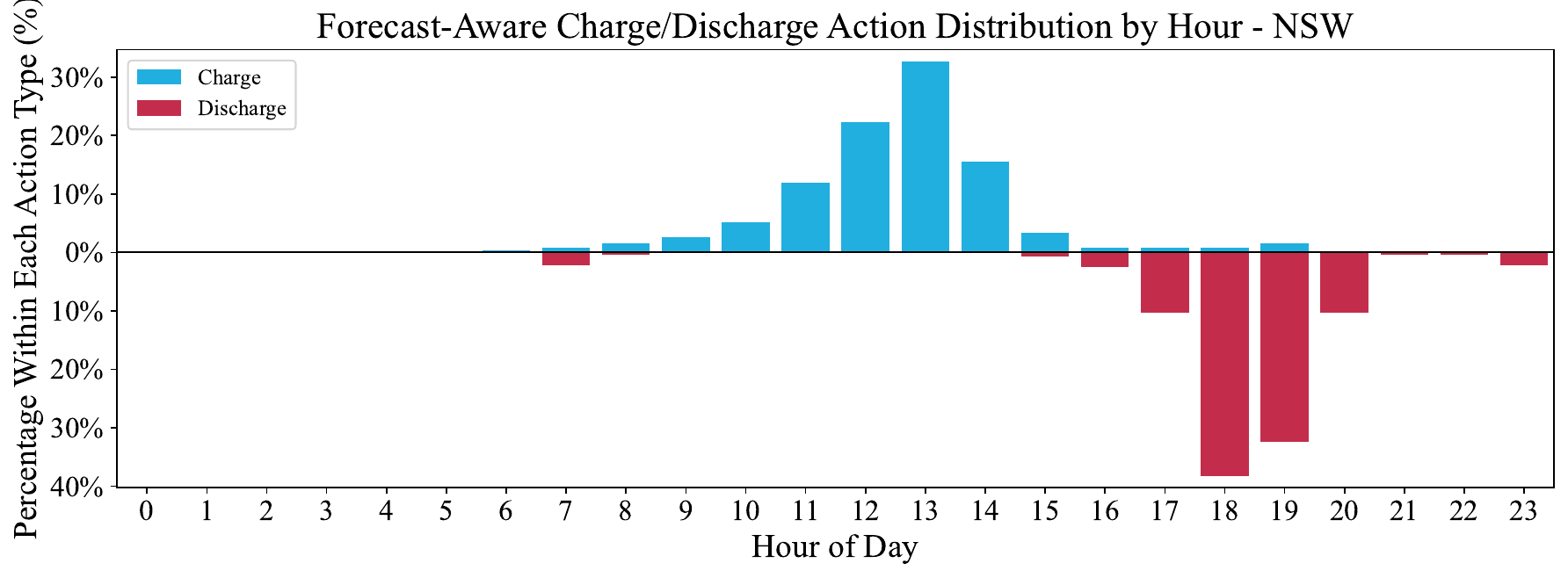}
        \caption{Forecast-Aware Algorithm}
        \label{fig:mid_histogram}
    \end{subfigure}
    
    \begin{subfigure}[b]{1\textwidth}
        \centering
        \includegraphics[width=\textwidth]{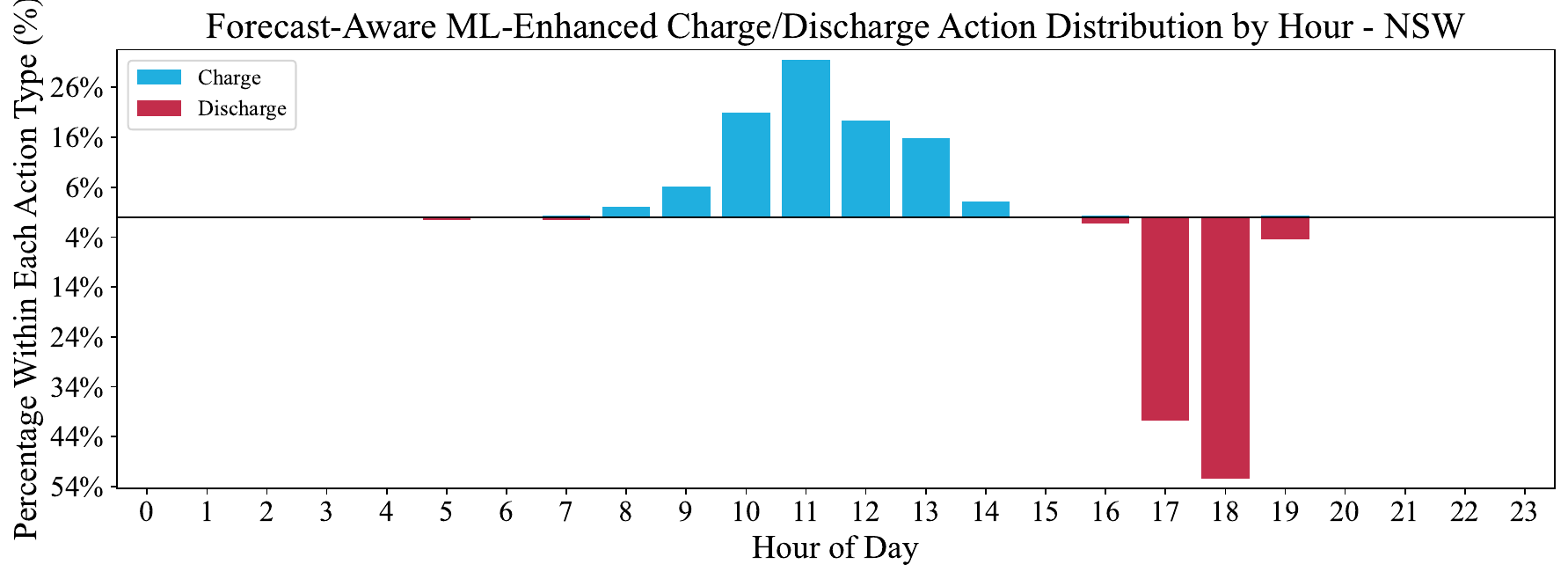}
        \caption{MILP-Optimised Algorithm with ML-Enhanced Price Forecasts}
        \label{fig:advanced_histogram}
    \end{subfigure}
    
    \caption{Comparison of Charge/Discharge Action Distributions by Algorithm}
    \label{fig:trading_histograms_combined}
\end{figure}

This section evaluates three BESS trading algorithms based on a revenue generation criterion. Each algorithm was tested on 2024 energy prices in the NSW region of the NEM. \autoref{fig:trading_histograms_combined} illustrates the evolution of trading behaviour across each algorithm.

\subsection{Forecast-Unaware Simple Trading Algorithm}
\label{sec:simple}

\begin{figure}[!ht]
    \centering
    \includegraphics[width=1\textwidth]{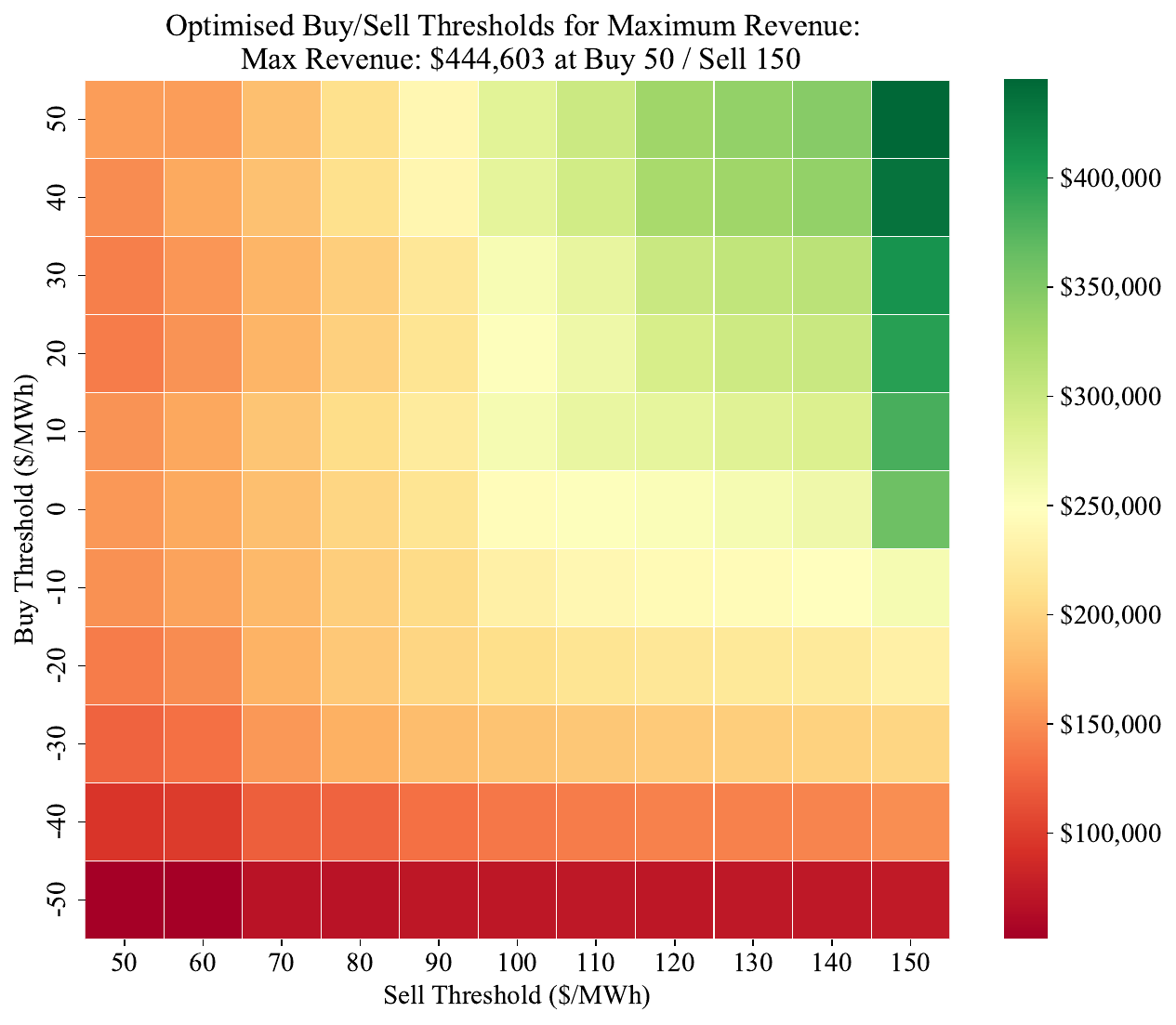}
    \caption{Optimised Buy/Sell Thresholds Sensitivity Analysis for Maximum Revenue}
    \label{fig:heatmap}
\end{figure}

\begin{figure}[!ht]
    \centering
    \includegraphics[width=1\textwidth]{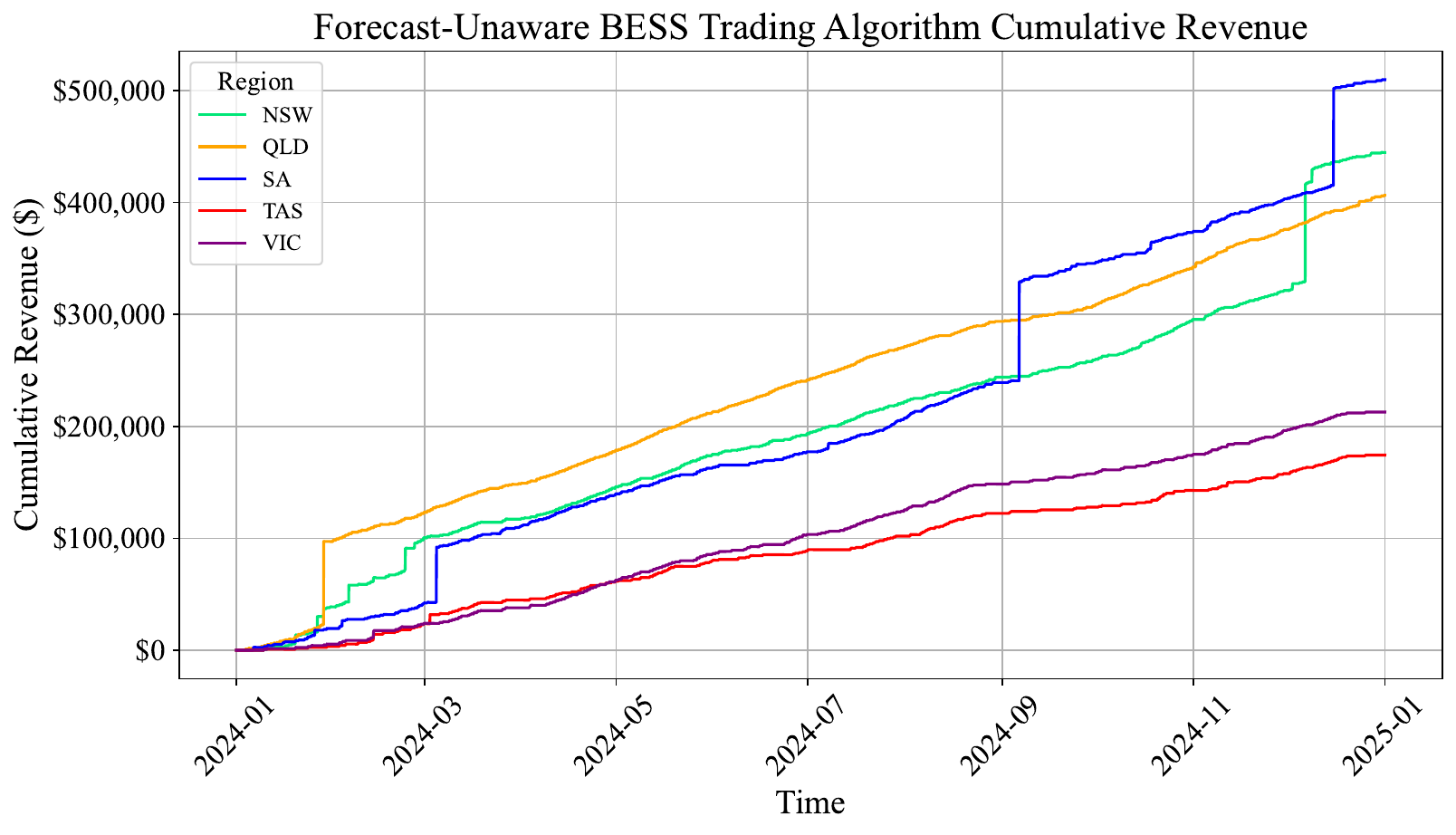}
    \caption{Forecast-Unaware BESS Trading Algorithm Cumulative Revenue}
    \label{fig:simple_revenue}
\end{figure}

The first stage of the trading algorithm development involved the implementation of a basic forecast-unaware strategy, with no foresight of market prices. The objective was to establish a benchmark for comparison with more advanced, forecast-informed models developed later in the study. This algorithm was designed to function solely on the SOC without reference to any predictive market data. The simplicity of the algorithm lay in its static decision-making rules, it would send AEMO a bid to buy energy and charge at a predetermined threshold when the battery's SOC was less than 50\% and it would likewise send AEMO an offer to sell energy and discharge at a predetermined threshold when the battery's SOC was at least 50\%. No adaptive learning or historical context was used in this decision-making process, and the thresholds themselves were set based on observed market price distributions and refined through a sensitivity analysis.
\\\\
To determine the optimal price thresholds for this baseline model, a sensitivity analysis was conducted across a wide range of buy and sell price combinations in NSW. This was visualised using the heatmap in \autoref{fig:heatmap}, which revealed that the most favourable outcome, within the constraints of this simple framework, occurred when the BESS was configured to buy electricity at or below \$50/MWh and sell at or above \$150/MWh. The resulting cumulative revenue of this threshold combination is displayed in \autoref{fig:simple_revenue} for all regions, and yielded total revenues ranging from \$180,000 to \$500,000 over the evaluation period. While this revenue provides a functional baseline, it is relatively modest, especially when considering the capital and operational costs typically associated with utility-scale battery systems. Nonetheless, the result is in line with expectations, given the rudimentary nature of the algorithm and its lack of responsiveness to dynamic market signals. The algorithm’s performance is inherently limited by its inability to anticipate market shifts or exploit price arbitrage opportunities that are less overt or transient in nature.

\vspace{20pt}

\begin{figure}[!ht]
    \centering
    \includegraphics[width=1\textwidth]{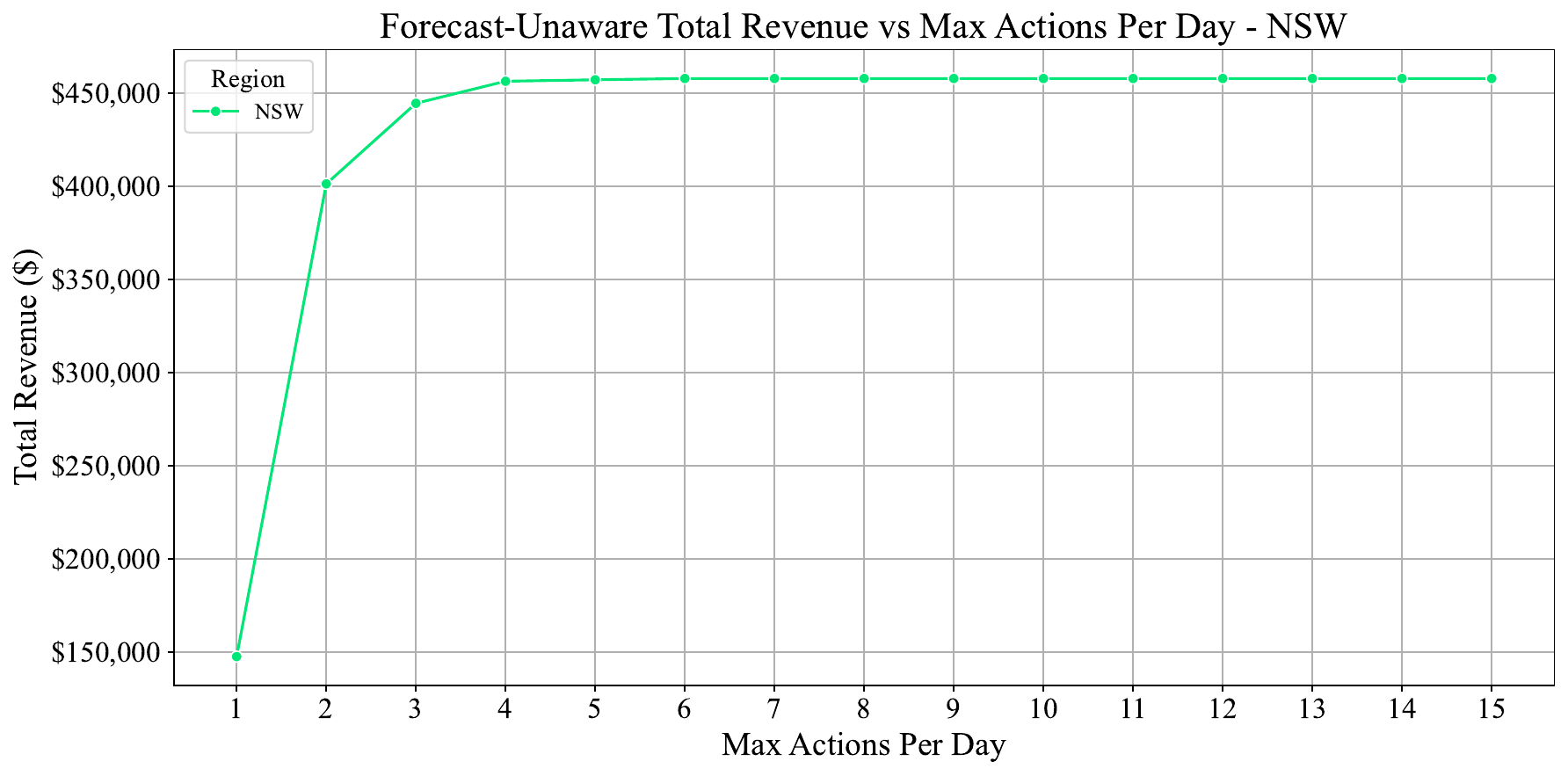}
    \caption{Forecast-Unaware Total Revenue vs Max Actions Per Day}
    \label{fig:Simple_Revenue_vs_MaxActionsPerDay}
\end{figure}

A particularly noteworthy observation is the pronounced revenue spike in \autoref{fig:simple_revenue} recorded for NSW in early December of 2024. Spikes like this can occur due to unexpected generation or infrastructure system failures, which can have wide reaching impacts on electricity markets \cite{spike_market_impact}. This surge can be attributed to an exceptionally high market price of \$17,480/MWh, which the algorithm successfully capitalised on. During this event, the BESS discharged 5 MWh of stored energy, resulting in a single-event revenue of approximately \$87,400 in 30 minutes. This instance highlights the capacity of even a simple, forecast-unaware algorithm to exploit extreme market volatility when the thresholds align with rare price surges, albeit in a largely opportunistic and infrequent manner. This infrequency is displayed in the result that over 2024, the total number of charge and discharge actions in NSW were 560, leading to 1.53 average actions per day, revealing that the algorithm is not fully utilising its maximum of three actions per day to generate revenue. More broadly, out of all the regions, these price spikes were only captured in NSW, QLD and SA, which can be explained by the high price volatility due to increased renewable penetration and more frequent contingencies due to the longer transmission network and harsher environment in northern Australia.
\\\\
Interestingly, the sensitivity analysis revealed a counterintuitive finding. When configured with highly restrictive and arguably unreasonable thresholds, with both buy and sell prices set at \$1,500/MWh, the algorithm achieved a significantly higher revenue, in the region of \$2 million for NSW. This sharp increase in revenue stems from the algorithm’s opportunistic exploitation of rare but extreme price spikes within the NEM. By waiting passively until such price peaks occurred, the BESS effectively operated as a peaking plant, discharging only when prices were exceptionally high. The profitability of this strategy was driven not by frequent activity or grid support but by the disproportionate financial gains associated with these infrequent yet dramatic market events.
\\
However, this mode of operation runs contrary to the underlying principles of BESS deployment in modern power systems. While the financial performance may appear attractive, a strategy that relies on sporadic market interventions fails to deliver the continuous balancing, smoothing, and reliability services that BESS assets are often intended to provide. Operating predominantly in reserve for rare events, the system remains largely idle, providing minimal support to grid stability or renewable integration efforts. Such a trading pattern reflects a fundamentally different role, akin to that of a traditional peaking plant, rather than a dynamic and responsive storage asset capable of enhancing market flexibility. In addition, when energy prices surge, there is a larger risk that higher BESS sell threshold prices will cause more expensive peaking plants to be pushed out of the market due to a lower clearing price being achieved, which would yield an overall lower revenue. Lastly, if a BESS is utilised as a peaking plant, it will not be able to provide optimal services in frequency response markets and private power purchase agreements, missing out on potentially lucrative revenue streams.
\\\\
This finding underscores a critical tension in the design of BESS trading strategies, the trade-off between maximising short-term financial returns and delivering broader system-level benefits. The simple forecast-unaware algorithm highlights this challenge vividly. While high revenues can theoretically be obtained through extreme thresholding, doing so may come at the expense of system reliability and broader policy goals. Therefore, while the baseline model offers a useful point of reference, its limitations reinforce the need for more nuanced, adaptive algorithms that balance profitability with system value. This consideration directly motivates the development of forecast-informed strategies explored in subsequent sections.
\\\\
To further evaluate the operational limitations of the simple forecast-unaware algorithm, an additional sensitivity analysis was conducted to examine how restricting the number of allowable trading actions per day influences total revenue. \autoref{fig:Simple_Revenue_vs_MaxActionsPerDay} plots total revenue as a function of the maximum number of trading actions per day in the NSW region. The results reveal a steep initial increase in revenue, with the total rising from approximately \$150,000 when the algorithm is limited to a single action per day, to around \$400,000 when allowed two actions. This sharp jump reflects the value of having the flexibility to both charge and discharge within a single day, thereby capturing intra-day price spreads more effectively. However, the curve begins to plateau beyond three to four actions per day, reaching a ceiling of around \$450,000. This flattening indicates diminishing marginal returns from increased activity, suggesting that most profitable opportunities are captured within just a few trades per day, and that energy prices do not fluctuate enough beyond the buy and sell thresholds to allow for more trades. Beyond this point, additional flexibility offers little to no revenue benefit, reinforcing the idea that a simple rule-based strategy can only extract limited value from market conditions, regardless of how often it is permitted to act.

\begin{figure}[!ht]
    \centering
    \includegraphics[width=1\textwidth]{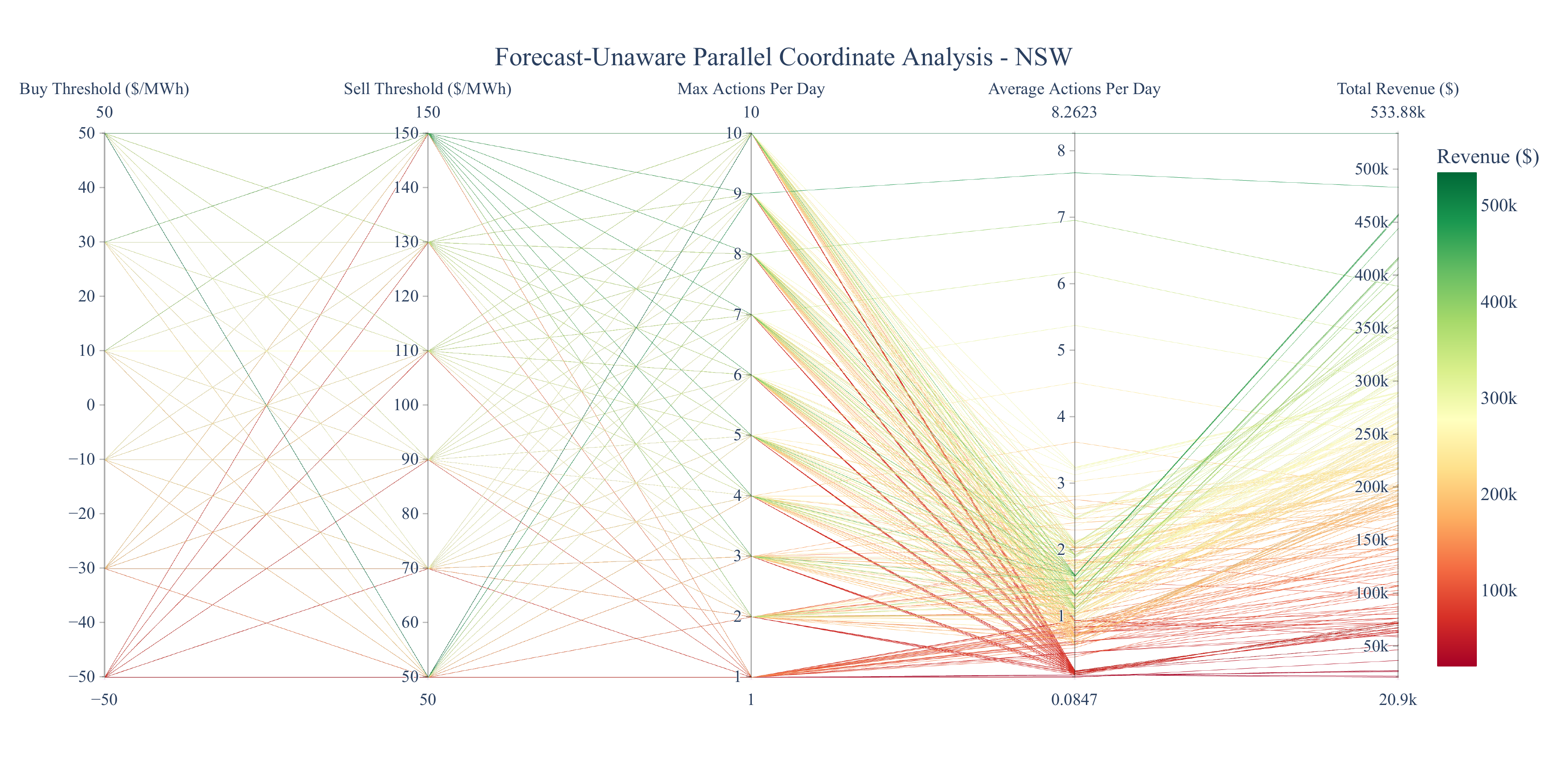}
    \caption{Forecast-Unaware Parallel Coordinate Analysis}
    \label{fig:Parallel_Coordinates_Revenue}
\end{figure}

To gain deeper insight into the behaviour of the simple forecast-unaware trading algorithm, \autoref{fig:simple_histogram} presents a histogram of trading activity distributed across each hour of the day. The most notable observation is the algorithm’s tendency to concentrate charge actions around 8:00 am and discharge actions around 5:00 am, with a marked reduction in activity during the intervening hours. This pattern corresponds with the typical morning and evening demand peaks within the NEM, a fascinating result suggesting that even without any forecast input, the algorithm naturally responds to underlying market dynamics. These periods also coincide with heightened price volatility and uncertainty, as previously demonstrated in \autoref{fig:NSW_accuracy}, further reinforcing the notion that the algorithm, despite its simplicity, is still indirectly influenced by broader demand-driven fluctuations in the market.
\\\\
To explore the complex interdependencies between key BESS operational parameters and both revenue generation and system utilisation, a parallel coordinate plot in \autoref{fig:Parallel_Coordinates_Revenue} was created using the full range of tested configurations. This visualisation allowed for simultaneous comparison of multiple variables, threshold settings, action constraints, utilisation rates, and resulting revenue, to offer a holistic perspective that is not easily captured through traditional one-variable-at-a-time sensitivity analyses. The plot revealed clear patterns that help inform optimal trade-offs between performance and long-term sustainability. Notably, the highest revenue was achieved when both the buy and sell thresholds were set at \$50/MWh and the algorithm was permitted to act up to nine or ten times per day. This configuration produced an average of 8.3 trading actions daily and yielded the highest financial return at over \$500,000. However, this level of activity is likely unsustainable in practice, as it would lead to rapid battery degradation, increased operational costs, and reduced overall asset lifespan. More realistically, the next most effective configuration involved a buy threshold of \$50/MWh, a sell threshold of \$150/MWh, and a maximum of between three and ten actions per day. This setup generated approximately \$450,000 in total revenue while requiring an average of just 1.7 actions per day, which is the same result found in \autoref{fig:simple_revenue}, and a far more sustainable operating profile for long-term BESS deployment. The parallel coordinate plot thus served not only as a diagnostic tool for revenue optimisation but also as a means of identifying balanced strategies that respect the physical and economic constraints of real-world battery systems.

\subsection{Forecast-Aware MILP-Optimised Trading Algorithm}
\label{sec:mid}

\begin{figure}[!ht]
    \centering
    \includegraphics[width=1\textwidth]{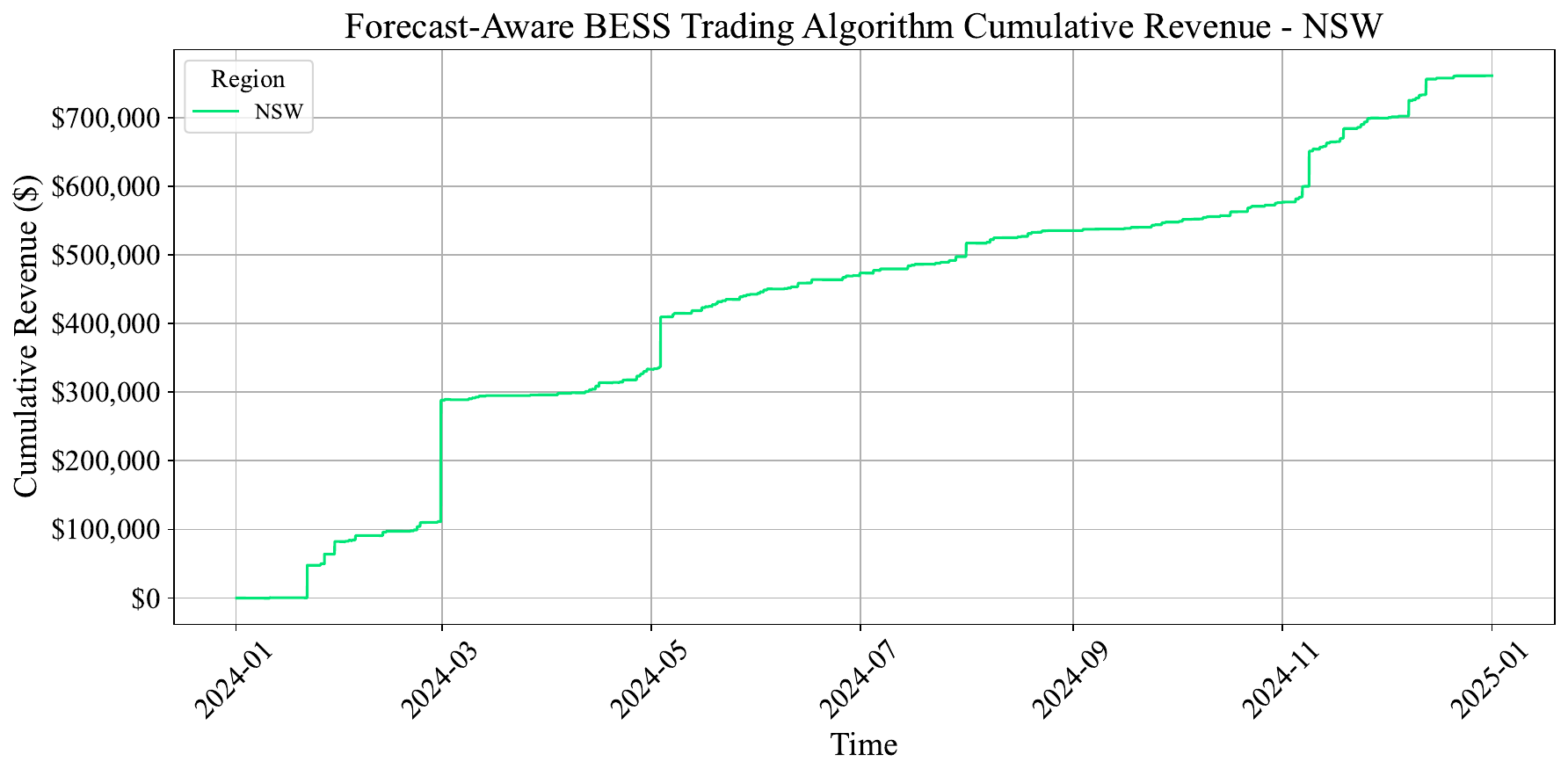}
    \caption{Forecast-Aware BESS Trading Algorithm Cumulative Revenue}
    \label{fig:mid_revenue}
\end{figure}

The forecast-aware MILP trading algorithm was designed to exploit the patterns and limitations of AEMO price forecast accuracy by strategically scheduling charge and discharge cycles based on forecasted prices. Unlike the baseline model, which operated independently of any forecast information, this algorithm directly referenced AEMO's 30-minute price forecasts to inform decision-making, with thresholds set to initiate a buy action when the forecasted price fell below \$50/MWh and a sell action when it exceeded \$150/MWh. These thresholds were not arbitrarily selected but were carried over from the optimal parameter region identified in the sensitivity analysis heatmap presented in Section \ref{sec:simple}. That analysis highlighted that these thresholds consistently produced one of the highest revenue outcomes, with \autoref{fig:mid_revenue} displaying the forecast-aware model achieving approximately \$750,000 in cumulative arbitrage revenue across the 2024 calendar year.
\\

Despite this stronger revenue figure, it is important to recognise that the algorithm's performance was fundamentally bounded by the inherent imperfections and uncertainties of the AEMO forecast data. While the algorithm's structure was flexible and capable of responding to profitable opportunities indicated by forecast price volatility, missed opportunities which stemmed from inaccurate forecast signals had a non-negligible impact on the overall result. Forecasts that either underestimated or overestimated price spikes or troughs led to suboptimal trading actions or, in some cases, inaction, thereby reducing potential revenue. Nevertheless, the algorithm’s responsiveness to forecast-informed signals still outperformed the naïve baseline strategy both in terms of annual revenue and trading efficiency.
\\\\
To maintain operational realism and prevent excessive cycling that could be economically or technically impractical for real-world BESS deployments, the model was constrained to a maximum of six trading actions per day to strike a balance between revenue generation and extend the operational lifespan of the BESS. Under this constraint, the algorithm executed a total of 546 trading actions over the course of 2024, averaging approximately 1.49 actions per day. This relatively conservative activity level suggests the algorithm was selective and chose to act only when forecast signals met the threshold criteria with sufficient confidence. Furthermore, the AEMO rejects any sell bids that are above the market clearing price, and vice versa. Hence, the low action rate was also observed due to the increased amount of rejected bids since the threshold sell price was relatively high at \$150/MWh.
\\\\
The \$750,000 revenue figure, while promising, can be viewed as a lower bound on the potential of forecast-informed trading, given the limitations of the AEMO forecasts. The inclusion of more accurate or supplementary forecasts such as those enhanced via weather forecasts could plausibly drive further gains. As it stands, however, the results demonstrate that a basic integration of publicly available forecast data, when processed with appropriate thresholds and operational constraints, can yield tangible financial benefits in a realistic market setting.
\\\\
To provide further insight into the behaviour of the forecast-aware trading algorithm, \autoref{fig:mid_histogram} presents a histogram showing the distribution of trading decisions across each hour of the day. In line with the forecast-unaware histogram displayed in \autoref{fig:simple_histogram}, the forecast-aware algorithm exhibits minimal trading activity between 9:00 pm and 6:00 am, reflecting limited arbitrage opportunities during these hours. However, less trades occurred early-on in the day, which can be explained by the algorithm being able to foresee future prices and postpone short term smaller gains for long term greater profitability. The majority of trades are concentrated around 1:00 pm for charge actions and again near 6:00 pm for discharge actions. Notably, there is a pronounced dip in trading frequency during the early afternoon which divulges the algorithm's selectivity in responding to forecast signals that meet the buy or sell thresholds with sufficient confidence. This temporal pattern highlights the algorithm’s tendency to align its actions with known daily price cycles while still being constrained by the limitations of forecast accuracy. Finally, the slight increase in discharge action frequency at 11:00 pm suggests that the MILP optimiser prioritises late trades to utilise any remaining daily actions and leftover stored energy before commencing the next day.

\vspace{30pt}

\subsection{MILP-Optimised Trading Algorithm with ML-Enhanced Price Forecasts}

\begin{figure}[!ht]
    \centering
    \begin{subfigure}[c]{0.495\textwidth}
        \includegraphics[width=\textwidth]{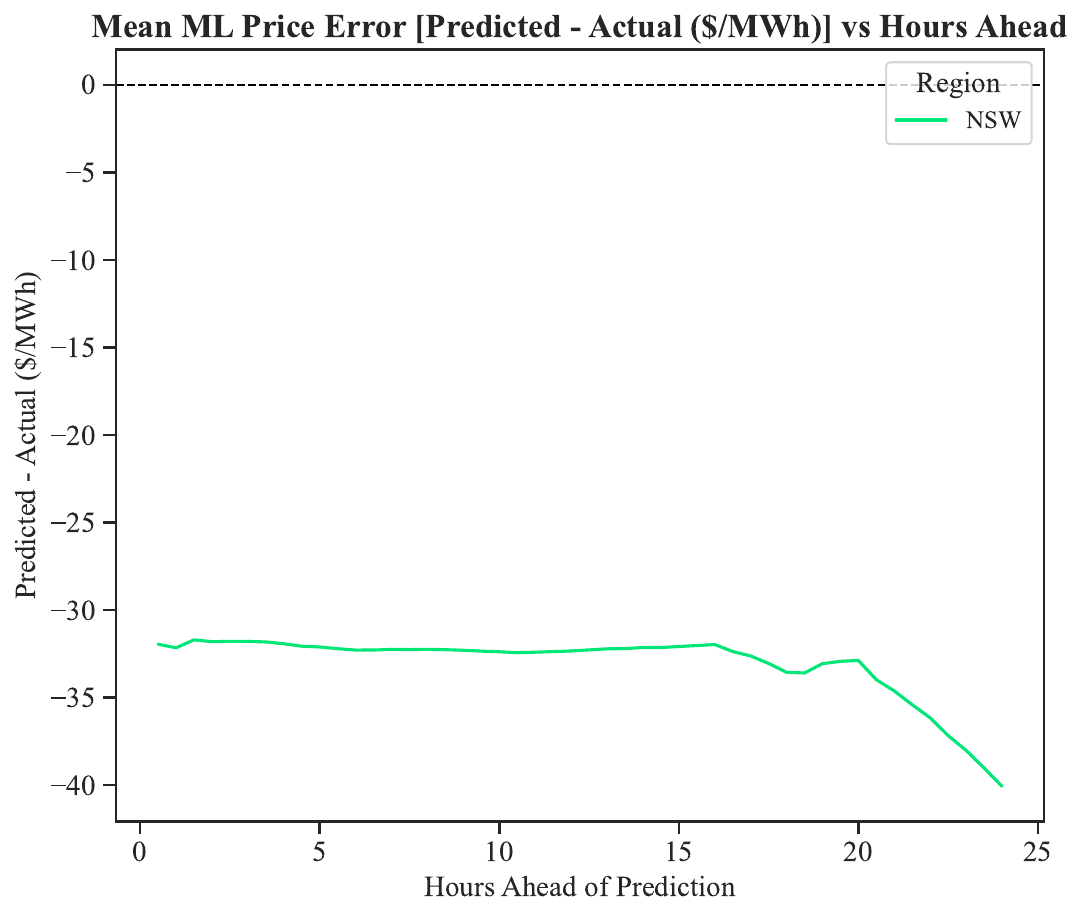}
        \caption{Mean Price Error}
        \label{fig:ML_mean_error}
    \end{subfigure}
    \hfill
    \begin{subfigure}[c]{0.495\textwidth}
        \includegraphics[width=\textwidth]{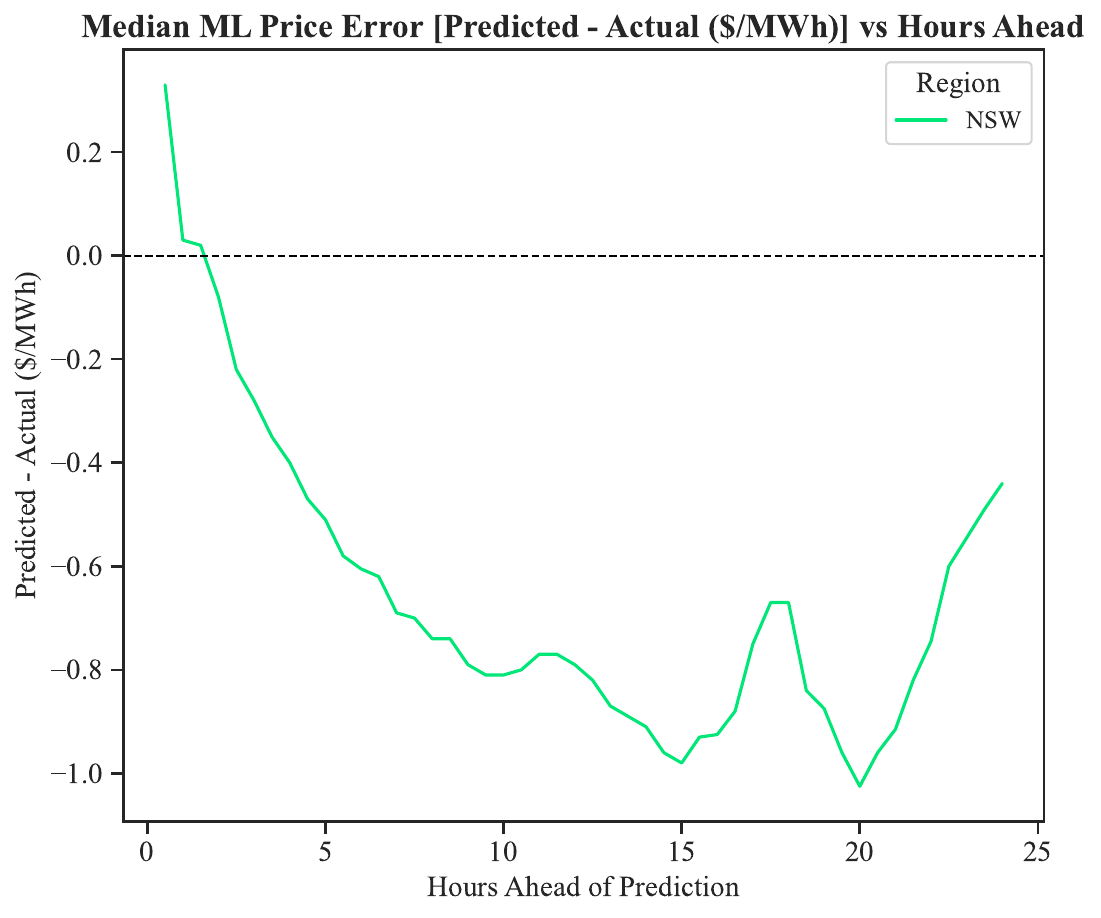}
        \caption{Median Price Error}
        \label{fig:ML_median_error}
    \end{subfigure}
    
    \caption{ML-Enhanced Forecast Regional Accuracy Ahead of Prediction}
    \label{fig:ML_Regional_Accuracy}
\end{figure}

\begin{figure}[!ht]
    \centering
    \includegraphics[width=1\textwidth]{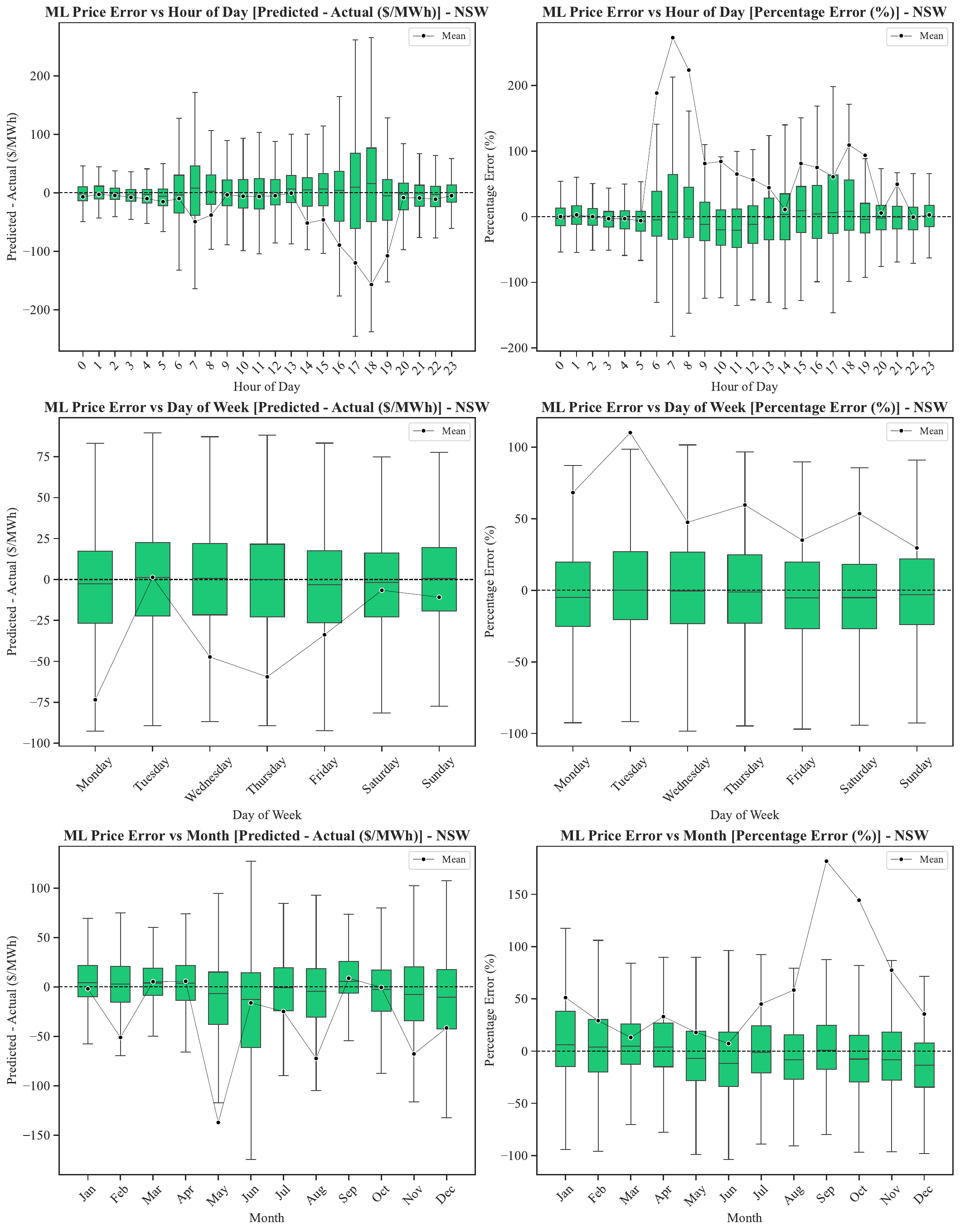}
    \caption{ML-Enhanced NSW Forecast Accuracy}
    \label{fig:ML_NSW_accuracy}
\end{figure}

The final and most advanced trading strategy developed in this study integrates machine learning-based price forecasting into a MILP optimisation framework. This model represents a significant evolution from the baseline and forecast-reliant strategies by using data-driven predictions to inform charge and discharge actions, rather than relying solely on the AEMO’s published forecasts. Crucially, the ML component is used exclusively to enhance the accuracy of price forecasts, and it does not influence or override the decision-making logic itself, which remains entirely governed by the MILP optimiser.
\\\\
The predictive engine at the core of this approach is a Random Forest Regressor implemented using the scikit-learn library. Hyperparameters in random forest models are configuration settings that define the structure and behaviour of the learning algorithm, rather than being learned from the data itself. Through a process of hyperparameter tuning via grid search and cross-validation, the model was optimised for the specific task of short-term energy price forecasting. The selected parameters: 100 estimators, a maximum tree depth of 10, a minimum samples split of 2, and a minimum samples per leaf of 2, strike a balance between model complexity and generalisation performance. This configuration was found to offer strong predictive power while maintaining stability across different market conditions and forecast horizons.
\\\\
The introduction of the ML-enhanced price forecast has led to a marked shift in error characteristics compared to the original AEMO forecasts. Notably, the mean price absolute error portrayed in \autoref{fig:ML_mean_error} now consistently underestimates energy prices by approximately –\$30/MWh for forecasts made between half an hour and 20 hours ahead. Beyond this window, the error begins to increase. However, \autoref{fig:ML_median_error} displays the median error, which tells a more stable story since values remain much closer to zero, beginning slightly positive, dipping to around –\$1/MWh at the 15-hour mark, then exhibiting a small bump between 15 and 20 hours as previously seen in \autoref{fig:Regional_Accuracy}, before rising again. Overall this suggests that the ML model has effectively dampened the prediction error ingrained in the AEMO forecast, however, improvements could still be made. When examining error by hour of day, day of week, and month, both absolute and percentage errors in \autoref{fig:ML_NSW_accuracy} depict dramatically reduced ranges, typically within –100 to 150 \$/MWh, a full order of magnitude lower than the AEMO forecasts. This clearly demonstrates that the ML model has internalised and corrected for the AEMO's tendency to overestimate prices, especially during peak periods. In congruence with \autoref{fig:NSW_accuracy}, error concentrations around 7:00 am and 6:00 pm on weekdays persist which highlights enduring challenges in forecasting volatile morning and evening demand peaks despite the improvements brought by machine learning. Interestingly, while the mean absolute error is generally negative, the percentage error is consistently positive. This likely reflects a systemic underestimation of low-price events, since even small underpredictions in absolute terms can translate to large percentage errors when actual prices are close to zero.

The price thresholds used in the decision logic, buying at prices below \$50/MWh and selling at prices above \$150/MWh, were derived from the sensitivity analysis heatmap discussed in Section \ref{sec:simple}. These thresholds had previously been identified as yielding optimal arbitrage revenue under forecast-driven strategies, and were therefore carried over into the machine learning model for consistency and comparability.

\vspace{30pt}

\begin{figure}[!ht]
    \centering
    \includegraphics[width=1\textwidth]{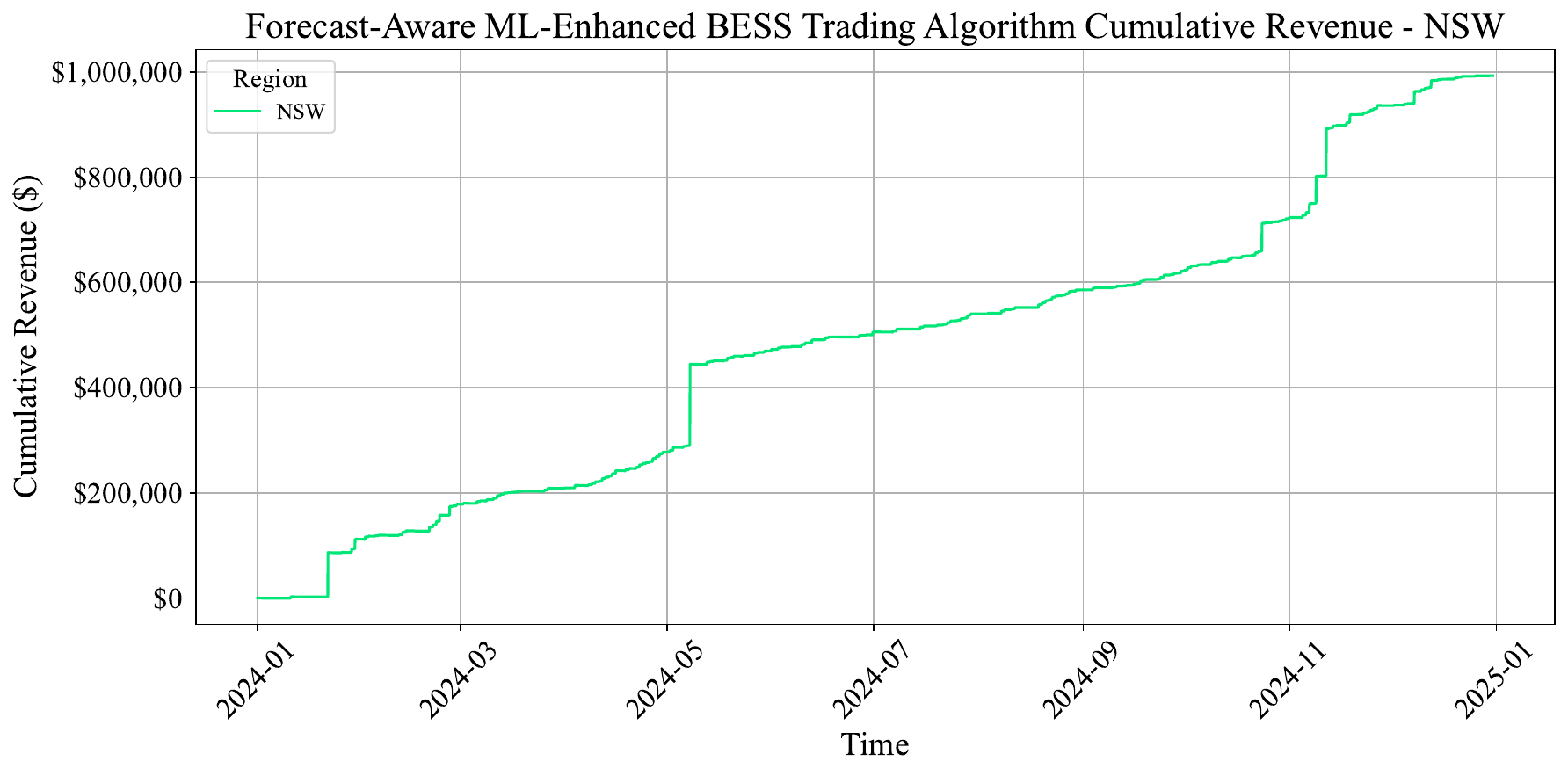}
    \caption{MILP-Optimised Algorithm with ML-Enhanced Price Forecasts Revenue}
    \label{fig:advanced_revenue}
\end{figure}

The MILP optimiser was constrained to execute a maximum of six charge or discharge actions per day, as in Section \ref{sec:mid}. Over the course of 2024, this led to a total of 762 charge or discharge actions, averaging approximately 2.08 actions per day. This algorithm generated approximately \$1 million in arbitrage revenue over the full year, as illustrated in \autoref{fig:advanced_revenue}. This result notably exceeds the revenue obtained from the AEMO forecast-based strategy and highlights the superior predictive accuracy of the Random Forest model in anticipating price peaks and troughs. By learning patterns in the AEMO forecast errors and capturing temporal and regional dynamics more effectively, the machine learning model was able to identify profitable opportunities that the AEMO forecasts alone did not expose.

Importantly, this approach does not entirely discard the value of AEMO forecasts but rather enhances them. Features derived from AEMO forecast values, including the future price trajectory and predicted deltas over multiple horizons, were used as key inputs to the machine learning model. This hybrid strategy allows the algorithm to correct for systematic forecast biases while still benefiting from AEMO’s domain-informed baseline expectations.

\vspace{50pt}

To further elucidate the behaviour of the forecast-aware ML-enhanced trading algorithm, \autoref{fig:advanced_histogram} presents a histogram of trading decisions distributed across each hour of the day. In contrast to the baseline and forecast-aware algorithms, which exhibited occasional trading activity between 8:00 pm and 8:00 am, the ML-based strategy entirely refrains from executing trades during this period. This suggests a more discerning approach to timing, likely informed by the model’s ability to anticipate less profitable conditions during off-peak hours. The algorithm instead prioritises the majority of its charging activity around 11:00 am and highly concentrates discharge actions in the early evening, with over 80\% of discharges occurring between 5:00 pm and 6:00 pm. A notable decline in activity is observed between these two intervals, indicating a targeted, high-confidence execution pattern that reflects both market dynamics and battery operation constraints.
\\\\
Overall, the results confirm that integrating ML-enhanced forecasts into a MILP-optimised BESS trading framework enables more effective exploitation of market dynamics than relying solely on AEMO forecasts. The model’s outperformance of the AEMO-based strategy highlights the added value of data-driven optimisation, especially when combined with adaptive and constraint-aware execution.

\section{BESS Financial Return Analysis}

Economic merit was assessed by discounting the cash‐flow streams of each trading strategy against a capital expenditure (\gls{CAPEX}) of AU\$8M, as specified in Section \ref{sec:data_sources}. Operating expenditure (\gls{OPEX}) was fixed at AU\$1 kWh/yr, equating to an annual charge of AU\$20,000, consistent with the levelised cost of storage (\gls{LCOS}) benchmarks published by Lazard \cite{Lazard_LCOS}. The OPEX was subtracted from each expected annual revenue found in Section \ref{sec:trading_performance}, to find the annual cashflow from years 1 to 20. All future cash flows were discounted over the typical BESS lifespan of 20 years \cite{BESS_lifespan} at a rate of 5.5\%, the central estimate of pre-tax weighted-average cost of capital (\gls{WACC}) adopted by AEMO’s Integrated System Plan \cite{Synergies_WACC}. The resulting NPV, IRR and payback periods are reported in \autoref{tab:NPV_IRR}.

\begin{table}[!h]
\centering
\caption{Summary of BESS Trading Algorithm Payback Periods, NPVs \& IRRs}
\label{tab:NPV_IRR}
\makebox[\textwidth]{%
\resizebox{1\textwidth}{!}{%
\begin{tabular}{l|ccc}
\toprule
 & \multicolumn{3}{c}{\textbf{Cash Flow (Income \textbf{--} Expenses, AU\$)}} \\
\multicolumn{1}{c|}{\textbf{Year}} & \textbf{Forecast-Unaware} & \textbf{Forecast-Aware} & \textbf{ML-Enhanced Forecast} \\
\midrule
\multicolumn{1}{c}{0} & -8M & -8M & -8M \\
\multicolumn{1}{c}{1\textbf{--}20} & 430k & 730k & 980k \\
\midrule
\textbf{Payback Period (years)} & 18.6 & 11 & 8.2 \\
\textbf{Discount Rate} & 5.5\% & 5.5\% & 5.5\% \\
\textbf{NPV} & -\$2.86M & \$0.72M & \$3.71M
 \\
\textbf{IRR} & 1\% & 7\% & 11\% \\
\bottomrule
\end{tabular}
}
}
\end{table}

The forecast-unaware algorithm performs poorly in financial terms. Although it generates a nominal annual surplus of approximately AU\$430k, the combination of high upfront CAPEX and modest operating margin leads to a payback period of 18.6 years, still within the typical 20-year operational lifespan of lithium-ion BESS assets \cite{BESS_lifespan}. When the cash flows are discounted, the project’s NPV deteriorates to -AU\$2.86M and the IRR barely reaches one per cent, well below the typical hurdle rate of 8-9\% \cite{hurdle_rate} required for energy storage projects. This result underscores the limitations of static, rule-based strategies that are blind to future price signals, as their performance is constrained by reactive, rather than proactive, behaviour.
\\\\
Introducing forecast awareness markedly improves the economics. By exploiting the time-varying reliability of AEMO price projections, annual earnings rise to AU\$730k and the simple payback compresses to 11 years, comfortably within the asset’s service life. The discounted cash‐flow profile turns positive with an NPV of AU\$0.72M and an IRR of 7\%, marginally above the assumed WACC but below hurdle rates. The use of forecast information not only enhanced the timing of charge and discharge events but also enabled more confident participation in volatile price periods, thereby boosting revenue.
\\\\
The forecast-aware ML-enhanced model delivers a step-change in value creation. Leveraging the Random Forest’s superior predictive accuracy within the linear-programming optimiser lifts annual arbitrage profit to AU\$980k after OPEX deductions. The payback period shortens to 8.2 years, effectively halving the capital recovery horizon relative to the baseline. Crucially, at the 5.5\% cent discount rate the project attains an NPV of AU\$3.71M, and an IRR of 11\%, clearing typical energy project hurdle rates of 8-9\% \cite{hurdle_rate}. Given that ancillary-service income and capacity market payments were deliberately excluded from this analysis, the ML-enhanced strategy can be viewed as a conservative yet compelling investment proposition.
\\\\
Overall, the results highlight the financial penalty of operating a BESS without foresight, the incremental benefit of incorporating market operator forecasts, and, most strikingly, the superior returns unlocked by a data-driven, forecast-aware optimisation framework. Under prevailing market and cost assumptions, only the ML-enhanced algorithm delivers both a sub-decennial payback and a double-digit IRR, firmly establishing it as a viable strategy for commercial deployment in the Australian NEM.

\begin{savequote}[95mm]
Then you will know the truth, and the truth will set you free.
\qauthor{\textbf{---}John 8:32}
\end{savequote}

\chapter{Discussion}
\label{ch: Discussion}

\minitoc


\section{Key Findings}

This research set out to investigate the value of forecast-informed optimisation in enhancing the financial performance of BESS participating in the Australian NEM. It did so by evaluating the accuracy of publicly available price forecasts issued by the AEMO, and by designing trading strategies of increasing sophistication. From a forecast-unaware baseline model, to a forecast-informed MILP optimiser, and finally to a machine learning–enhanced model capable of dynamically correcting forecast errors before feeding future price predictions into the MILP optimiser. The findings present a strong empirical case for the integration of forecast reliability into BESS dispatch strategies, whilst also exposing several structural features of market forecasts that are of broader relevance to energy system modelling and algorithmic market participation.
\\\\
A primary empirical finding of the study is that AEMO's energy price forecasts exhibit a highly structured error distribution. Forecast accuracy diminishes with longer horizons in a broadly predictable fashion, though a sharp discontinuity in error behaviour is observed beyond approximately 30 hours. This inflection point is hypothesised to reflect a transition in AEMO's forecasting methodology, likely moving from short-term dynamic models to medium-term statistical or historical pattern extrapolations. Such a regime shift in forecast composition has profound implications for the design of trading strategies. Short-horizon forecasts can be used directly within optimisation frameworks, whereas longer-term forecasts may require either correction or supplementation through adaptive predictive tools, such as machine learning. Furthermore, the consistent bias in AEMO forecasts, namely a tendency to overestimate prices during certain hours and seasons, provides an exploitable asymmetry that, if correctly modelled, enhances arbitrage decision-making and thus profitability of BESS development projects.
\\\\
In terms of regional analysis, the study finds substantial spatial variation in both market volatility and forecast reliability. QLD and SA demonstrate elevated forecast errors and high volatility, likely due to their significant penetration of variable renewable energy and vulnerability to transmission constraints and localised demand spikes. By contrast, NSW offers a more favourable combination, as it displays considerable price volatility, thereby providing arbitrage opportunities, whilst maintaining forecast error within manageable bounds. The decision to restrict the algorithmic simulations to NSW was therefore not arbitrary, but informed by an empirical optimisation of market conditions. This observation invites broader reflection on the role of spatial analytics in BESS deployment. While much of the discourse on storage profitability has focused on temporal arbitrage, the results here suggest that spatial forecast reliability and region-specific network constraints all factor critically into the realisable revenue of storage assets.

\vspace{20pt}

Another key implication of the research relates to the capacity of BESS algorithms to exploit recurring market rhythms. Both the baseline and advanced models tend to favour certain hours of the day, notably early morning and evening peaks, reflecting the cyclic structure of electricity demand and price formation in the NEM. However, the machine learning–augmented forecast algorithm exhibited a more refined and selective trading profile, actively avoiding low-value or noisy hours, and focusing its activity on price extremities with greater forecast confidence. This suggests that while even simple algorithms can passively benefit from daily cycles, truly optimised trading depends on the nuanced integration of time-sensitive features and probabilistic forecasting into the decision architecture.
\\\\
The financial outcomes further corroborate the added value of forecast-aware optimisation. The baseline threshold-driven model produced modest revenues, primarily capitalising on rare price spikes and revealing the inherent limitations of static rules in a dynamic market. The MILP-optimised forecast-aware model achieved significantly higher revenue by incorporating predictive insight and dynamically scheduling charge-discharge cycles over a rolling planning horizon. Most notably, the machine learning–enhanced model generated the highest cumulative revenue, despite operating under the same operational constraints. By dynamically forecasting on recent data and modelling the temporal and cyclical structures of the market, it consistently outperformed the model reliant on AEMO forecasts alone. The magnitude of improvement, an approximate 120\% increase in revenue over the baseline model, underscores the commercial value of data-driven trading intelligence in an increasingly volatile electricity landscape.
\\\\
At a systemic level, the findings of this study reaffirm that BESS profitability is not solely a function of price volatility or capacity, but is fundamentally determined by the intelligence and responsiveness of its dispatch strategy. This conclusion is consistent with current research showing that storage systems operating as intelligent agents, responsive to market conditions and forecast uncertainty, are significantly more profitable than those governed by static heuristics \cite{FANTASTIC_AEMO_FORECAST_PAPER, Arbitrage_Under_Price_Uncertainty, using_LP_optimisation_in_BESS}. The superior performance of the MILP optimiser and machine learning–enhanced models in this study validates this premise and reveals that even publicly available AEMO forecasts, when systematically exploited, can yield substantial returns, with pleasant NPV and IRR financial metrics outweighing typical hurdle rates. These results align with recent work in the Australian context that highlights the growing importance of forecast-driven scheduling in markets with high renewable penetration and increasing volatility \cite{probabilistic_NEM_forecast, FANTASTIC_AEMO_FORECAST_PAPER}. Moreover, the integration of machine learning to correct forecast biases supports the probabilistic modelling approaches advocated in studies such as \cite{probabilistic_NEM_forecast, proprietary_algorithms}, where volatility-aware forecasting was shown to improve dispatch efficiency. Taken together, the evidence suggests that battery operators who invest in adaptive optimisation, especially those capable of leveraging forecast structure and error dynamics, are better positioned to extract value in markets such as the NEM where imperfect foresight is the norm but not a barrier to profitability.

\section{Limitations and Challenges}

Notwithstanding the promising findings, several limitations of the study must be acknowledged, each of which bears directly on the interpretation of the results and the generalisability of the proposed algorithms.
\\\\
Most significantly, the modelling framework is confined exclusively to energy arbitrage in the spot market, without consideration of ancillary service participation. In the Australian NEM, BESS assets are increasingly active not only in energy arbitrage but also in the provision of frequency control ancillary services (\gls{FCAS}), inertia support, system strength, and reserve capacity, with FCAS typically representing the most lucrative revenue stream \cite{USEFUL_TO_MODEL_BESS_ARBITRAGE_AND_FCAS}. These services often offer substantial revenue streams, particularly under periods of market stress or contingency events. The exclusion of such services in the present study means that the reported revenues represent a conservative lower bound on the true revenue potential of a real-world BESS system.
\\
Moreover, the study does not account for capacity payments or flexibility incentives, both of which are being actively considered in market reforms in jurisdictions facing increasing penetrations of renewable energy. These mechanisms are designed to reward resources such as BESS that can provide rapid, dispatchable capacity or respond flexibly to supply-demand imbalances. Were such payments included in the model, the profitability and payback period of the storage asset would likely improve markedly. Consequently, the exclusive reliance on energy arbitrage understates the commercial viability of BESS in integrated markets and fails to capture the full scope of value streams available to BESS owners.
\\\\
Another challenge lies in the modelling assumptions around battery degradation and operational constraints. While the study incorporates basic degradation logic, applying a uniform 0.005\% reduction in capacity per cycle, this simplification does not fully reflect the complex chemistry- and temperature-dependent nature of battery wear. In reality, factors such as cell depth of discharge, rate of charging, ambient temperature, and resting SOC all influence degradation trajectories. More granular degradation models, particularly those parameterised for specific technologies such as lithium-iron phosphate or nickel-manganese-cobalt batteries, would allow for more accurate simulations of lifetime profitability and maintenance scheduling \cite{BESS_advanced_degradation}.
\\\\
Further, the study adopts a fixed configuration of 10 MW/20 MWh and assumes constant access to perfect execution at forecast-derived bid prices. In practice, BESS operators face constraints in terms of market access, bidding behaviour, regulation compliance, and occasionally, curtailment or grid congestion. These real-world frictions may reduce the frequency or profitability of optimal trades, even if correctly identified by the optimiser. Moreover, the MILP optimiser assumes convex cost and revenue structures and makes deterministic decisions based on point forecasts. It does not currently incorporate stochastic optimisation or scenario-based risk mitigation, which may be required in highly uncertain or adversarial markets.
\\\\
Finally, while the machine learning model demonstrably improves forecast precision and trading outcomes, it is subject to the typical limitations of supervised learning in non-stationary environments. Market conditions evolve, policy changes may restructure price formation, and unforeseen externalities, such as weather events or geopolitical shocks can induce distributional shifts in the data. More frequent retraining of the model would partially mitigate this issue, but more robust solutions, such as ensemble learning, online learning, or transfer learning across regions may offer improved generalisation in future implementations.

\section{Potential for Real-World Implementation}

The results of this study offer a robust theoretical foundation for the real-world deployment of intelligent BESS trading strategies. However, translating these findings from simulation to operation requires careful consideration of technical, regulatory, and economic realities.
\\\\
From a technical perspective, the core components of the forecast-aware and machine learning–enhanced algorithms are implementable using contemporary computing infrastructure. The use of Python-based optimisation frameworks, such as Gurobi, and well-established machine learning libraries ensures that the methods are reproducible and extensible. Cloud-based or edge computing environments can host the algorithms in near-real time, with data pipelines drawing continuously from AEMO’s publicly available forecast feeds and market pricing data. The rolling horizon structure of the optimiser and predictive model aligns well with operational best practices in the energy trading field.
\\\\
However, real-world implementation would necessitate a robust system for exception handling, bid rejection monitoring, and adaptive recalibration in the face of unexpected events. The latency and reliability of data inputs, including delays in forecast publication or communication failures, would need to be addressed through fault-tolerant design. In practice, execution risk, where recommended trades are not successfully placed or cleared, would need to be incorporated into the decision framework, perhaps through probabilistic bid acceptance modelling or real-time feedback loops.
\\\\
Regulatory approval and compliance also pose challenges. In the NEM, bidding strategies must adhere to strict transparency, dispatchability, and fairness guidelines, particularly for semi-scheduled assets. Algorithmic strategies that are opaque or that could be construed as gaming the market may attract scrutiny. To mitigate this, any implementation must be fully auditable and capable of generating explanatory logs of decision rationale. Furthermore, the system must remain compatible with evolving market rules, such as five-minute settlement reform, revised FCAS structures, and potential real-time reserve markets. As highlighted by recent research, the frequency and scale of rebidding, particularly by BESS using automated bidding systems, has increased substantially in recent years, contributing to unpredictable price swings and forecast errors \cite{FANTASTIC_AEMO_FORECAST_PAPER}. This growing volatility undermines effective scheduling and threatens market integrity which underscores the need for stricter oversight and potential reform of rebidding provisions to preserve transparency and stability in a high-storage, high-renewables future.
\\\\
Market participants should not underestimate the seriousness of complying with the National Electricity Rules (\gls{NER}). The \$900,000 penalty imposed on Neoen's Hornsdale Power Reserve by the Federal Court serves as a cautionary example \cite{Neoen_Fuckup}. Between July and November 2019, Neoen offered contingency FCAS it could not reliably deliver, breaching the NER despite receiving payment. This incident underscores the need for BESS operators to ensure that all market offers are both technically feasible and transparently justified. Non-compliance not only attracts significant legal and financial consequences but also undermines trust in the integrity of emerging energy technologies within the NEM.

\vspace{30pt}

On the economic front, the payback period of a BESS investment remains a critical determinant of its attractiveness. Considering that the algorithms developed in this study achieved substantial annualised revenues while operating exclusively within a single market and relying solely on energy arbitrage, the results are highly promising. They suggest that even without engaging in the complexity of multi-market participation, a BESS investment could be distinctly profitable. This is particularly noteworthy given the high capital cost of utility-scale storage, which often exceeds AU\$8 million for a 10 MW/20 MWh system. The ability to generate strong returns through a forecast-informed arbitrage strategy alone indicates that sophisticated, multi-layered optimisation may not be a prerequisite for financial viability. However, when ancillary services, capacity payments, and regulatory incentives are layered into the revenue stack, the payback horizon shortens considerably. This suggests that asset managers who are able to aggregate and monetise multiple value streams, through co-optimisation platforms or aggregator business models are best positioned to benefit from intelligent trading strategies.
\\\\
The broader system-level implications are equally noteworthy. If deployed at scale, forecast-informed BESS can do more than capitalise on market inefficiencies, they can reduce price volatility, improve grid stability, and facilitate renewable integration. By operating dynamically across energy and ancillary service markets, these systems can act like miniature distributed system operators which respond to supply-demand imbalances and offset the intermittency of wind and solar. As such, the algorithms developed in this study will contribute not only to private profitability but also to public system benefit.
\\\\
In summary, while real-world implementation entails technical and regulatory challenges, the forecast-informed trading strategies developed in this thesis are both feasible and economically attractive. They offer a pathway to more intelligent, responsive, and profitable energy storage operations, critical as energy systems evolve toward greater renewable penetration and increasingly dynamic markets.

\begin{savequote}[\textwidth/2]
``Begin at the beginning,'' the King said gravely, ``and go on till you come to the end: then stop.''
\qauthor{\textbf{---}Lewis Carroll, \textit{Alice in Wonderland}}
\end{savequote}

\chapter{Conclusion} 
\label{ch: Conclusion}

\minitoc


\section{Summary of Research Contributions}

This thesis has provided a comprehensive and data-driven investigation into the feasibility and financial performance of forecast-informed BESS trading strategies in the Australian NEM. Through a structured analysis that combined empirical forecast error evaluation, algorithmic trading simulation, and machine learning–driven enhancement, the research delivers a novel and rigorous framework for understanding how the accuracy of publicly available electricity price forecasts can be operationalised in real-world energy trading.
\\\\
The first major contribution lies in the characterisation of AEMO forecast reliability across multiple dimensions. By evaluating both absolute and percentage forecast errors across temporal and regional axes, the study revealed critical insights into when and where AEMO forecasts can be considered most dependable. The identification of a forecast accuracy discontinuity at approximately 30 hours ahead provides empirical evidence for a methodological shift in AEMO’s forecasting pipeline. This structural breakpoint had not been previously documented in the literature and offers a practical threshold for algorithm design. Further, the analysis of regional volatility and forecast error variance led to the selection of NSW as the optimal region for BESS deployment, due to its favourable combination of high volatility and relatively predictable forecast behaviour.
\\\\
The second substantive contribution is the development and performance evaluation of three increasingly sophisticated BESS trading algorithms: a simple threshold-based strategy that relies solely on current market prices, a MILP optimiser informed by AEMO forecasts, and an enhanced strategy that integrates machine learning forecasts into the MILP optimisation framework. The comparison of these models over the 2024 trading year in NSW demonstrated a clear and significant uplift in revenue as forecasting intelligence increased. The ML-enhanced strategy achieved the highest revenue, illustrating the value of predictive analytics and dynamic learning in operational energy trading.
\\\\
Finally, the thesis contributes a methodology for combining real-time forecast reliability with constrained optimisation. This architecture includes degradation-aware constraints, realistic action limits, and a rolling-horizon approach that simulates imperfect foresight. By incorporating these elements, the framework reflects the practical complexities of BESS operation in wholesale energy markets, thereby extending beyond stylised academic models toward a deployable system architecture.

\section{Practical and Theoretical Implications}

From a practical standpoint, the findings offer compelling evidence that intelligent trading strategies grounded in forecast evaluation can materially improve the economic viability of BESS assets. The magnitude of improvement observed, up to 120\% increases in annual revenue compared to naïve trading strategies, has direct implications for asset financing, risk management, and return on investment for energy storage developers and operators. These improvements were achieved without assuming privileged access to data or exclusive market insights, but rather by using publicly available AEMO forecasts and scalable algorithmic tools.
\\\\
One particularly salient practical implication concerns the alignment between battery operation and market structure. The results indicate that arbitrage strategies informed by forecast prices can generate consistent and meaningful revenue without necessitating extreme trading behaviour such as a BESS operating as a peaking plant seen in Section \ref{sec:simple}. This stands in contrast to simplistic strategies that chase rare price spikes and thereby risk underutilising storage assets.
\\\\
Theoretically, the study underscores the importance of uncertainty quantification in energy market modelling. Forecasts are not binary indicators of future prices but probabilistic signals whose utility depends on context-specific reliability. By showing that forecast errors are not uniformly distributed, either temporally or spatially, this thesis argues for the development of adaptive trading models that respond to known limitations of predictive data. This reframing of forecasts as fallible but strategically useful inputs contributes to a broader theoretical understanding of bounded rationality in automated energy trading systems.
\\\\
Moreover, the research highlights the relevance of machine learning as a forecasting enhancement tool, not a replacement for market fundamentals. The most effective strategy in this study did not discard AEMO forecasts but used them as features to be refined through data-driven models. This hybrid approach provides a theoretically grounded compromise between deterministic modelling and fully autonomous black-box decision-making which suggests that future systems will need to combine domain knowledge with statistical learning to achieve optimal performance.

\section{Recommendations for Industry and Policy}

For industry stakeholders, the findings of this thesis advocate for the adoption of forecast-aware trading frameworks in the operation of BESS assets. Specifically, market operators should invest in predictive analytics capabilities that can evaluate, correct, and supplement publicly available forecasts. The added revenue from such systems is not only financially meaningful but can also de-risk investments in energy storage, especially in markets characterised by volatility and low forward contract liquidity. Additionally, developers may consider siting decisions not only on raw price spreads but also on forecast reliability metrics, since profitability hinges on the ability to anticipate, not merely respond, to price movements.
\\\\
Battery integrators and technology vendors are likewise encouraged to embed forecast-aware optimisation modules into their control systems. These modules should not only optimise for price arbitrage but also incorporate operational constraints, degradation models, and market rule compliance features to ensure robust, real-world deployability. In doing so, BESS can shift from passive grid participants to active market agents capable of extracting and providing value in a coordinated, intelligent manner.
\\\\
From a policy perspective, the results call attention to the latent economic potential of improved forecasting infrastructure. AEMO's existing price forecasts, whilst sufficient for broad market participation, contain systematic errors that constrain their utility for short-term, high-resolution trading. Regulatory bodies may consider incentivising the development of higher-resolution, probabilistic forecasts, or enabling third-party forecast providers to contribute to market transparency. Policies that reward BESS assets for system-supportive behaviours, such as peak shaving or congestion relief, should also recognise the role that intelligent dispatch plays in achieving those outcomes.

\vspace{20pt}

Importantly, the thesis also identifies a gap in current market design, the absence of mechanisms to reward trading strategies that promote grid stability rather than profit from peak prices. Without such mechanisms, highly selective spike-chasing behaviour may disincentivise consistent participation and fail to harness the full value of BESS flexibility. Policymakers should explore hybrid incentive structures that align private profit motives with public system reliability goals.

\section{Future Research Directions}

While this thesis advances the state of knowledge in BESS trading optimisation, it also opens several avenues for future inquiry. First, the scope of this study was limited to energy arbitrage in the spot market. In reality, BESS assets often participate concurrently in ancillary services markets, such as FCAS, reserve provision, and, in some jurisdictions, inertia or voltage control markets. A natural extension of this work would be to develop a co-optimisation framework that simultaneously schedules energy trading and ancillary service participation, thereby reflecting the full revenue stack available to storage assets.
\\\\
Second, the models presented here treat forecasts as point estimates. Future research could incorporate uncertainty modelling by leveraging probabilistic or scenario-based forecasts. Such models would enable risk-adjusted optimisation, allowing storage systems to trade not only based on expected value but also on confidence intervals or tail-risk probabilities. Bayesian learning, ensemble forecasting, and Monte Carlo–based approaches are promising candidates for this extension.
\\\\
Third, the study uses a static degradation model and simplified operational constraints. More advanced models simulating electrochemical and thermal battery behaviour could improve long-term profitability estimates and inform warranty compliance and maintenance scheduling. They could also simulate heterogeneous BESS fleets with varying technical characteristics to enable more granular market-wide analyses.
\\
Fourth, while the present research focused on the Australian NEM, the methodological framework is generalisable. Applying the algorithmic architecture to other markets, such as PJM and ERCOT in the US, or EPEX in Europe, would test its robustness across regulatory regimes and grid structures. Cross-market comparisons could also reveal which forecasting practices or market rules most effectively support intelligent BESS integration.
\\\\
Finally, future work could explore agent-based simulations of storage participation in competitive environments. As more BESS systems enter the market with forecast-informed algorithms, their collective behaviour could lead to price dampening, volatility suppression, or emergent strategic interactions. Modelling these dynamics would provide insight into market saturation thresholds, revenue cannibalisation risks, and the systemic implications of widespread storage deployment. In addition, future research could also incorporate machine learning directly into the trading decision-making, as opposed to the MILP optimisation performed in this study. Reinforcement learning methods such as Q-learning and advantage actor-critic can optimise decisions under uncertainty and adapt to forecast errors, potentially outperforming rule-based approaches. These techniques have shown promise in electricity market trading \cite{future_deep_reinforcement, future_advantage_actor_critic}.
\\\\
In conclusion, this work demonstrates that forecast-informed, algorithmically optimised BESS trading strategies offer substantial improvements over naïve or static approaches. By combining empirical forecast evaluation, adaptive optimisation, and machine learning, the study charts a path toward more intelligent, efficient, and profitable energy storage systems, capable not only of arbitrage but of contributing to the broader goals of energy transition, decarbonisation, and market resilience.



\startappendices
\chapter{Trading Algorithms}
\label{app:endix}

\begin{algorithm}[!ht]
\caption{Threshold-Based Forecast-Unaware BESS Trading Strategy}\label{alg:simple}
\begin{algorithmic}[1]
\Require $P_{\text{charge}},~P_{\text{discharge}},~C_{\text{init}},~C_{\text{current}},~\eta,~R,~A_{\text{max}},~\Delta t$
\LeftComment{ \textit{$P_{\text{charge}}$, $P_{\text{discharge}}$: threshold prices}}
\LeftComment{ \textit{$C_{\text{init}}$: initial battery capacity (MWh)}}
\LeftComment{ \textit{$C_{\text{current}}$: current SOC (MWh)}}
\LeftComment{ \textit{$\eta$: battery degradation rate}}
\LeftComment{ \textit{$R$: total revenue}}
\LeftComment{ \textit{$A_{\text{max}}$: max daily actions}}
\LeftComment{ \textit{$\Delta t$: settlement period}}

\Procedure{SimpleBidStrategy}{$\text{MarketPrice}[t],~\text{Date}[t]$}
\State Initialise $C_{\text{max}} \gets C_{\text{init}},~C_{\text{current}} \gets 0.5 \cdot C_{\text{init}}$
\State Initialise $\text{DailyActionCount}[\cdot] \gets 0$
\For{$t \in [1,~T]$}
    \State $p \gets \text{MarketPrice}[t]$
    \State $d \gets \text{Date}[t]$
    \If{$\text{DailyActionCount}[d] \geq A_{\text{max}}$}
        \State \textbf{continue}
    \EndIf
    \State $P_{\text{bid}} \gets \begin{cases}
        P_{\text{charge}}, & \text{if } C_{\text{current}} < 0.5 \cdot C_{\text{max}} \\
        P_{\text{discharge}}, & \text{otherwise}
    \end{cases}$

    \If{$P_{\text{bid}} = P_{\text{charge}}~\textbf{and}~P_{\text{bid}} \geq p~\textbf{and}~C_{\text{current}} < C_{\text{max}}$}
        \State $E \gets \min(P_{\text{max}} \cdot \Delta t,~C_{\text{max}} - C_{\text{current}})$
        \State $R \gets R - E \cdot p$
        \State $C_{\text{current}} \gets C_{\text{current}} + E$
        \State $C_{\text{max}} \gets C_{\text{max}} \cdot (1 - \eta)$
        \State $\text{DailyActionCount}[d] \gets \text{DailyActionCount}[d] + 1$
        
    \ElsIf{$P_{\text{bid}} = P_{\text{discharge}}~\textbf{and}~P_{\text{bid}} \leq p~\textbf{and}~C_{\text{current}} > 0$}
        \State $E \gets \min(P_{\text{max}} \cdot \Delta t,~C_{\text{current}})$
        \State $R \gets R + E \cdot p$
        \State $C_{\text{current}} \gets C_{\text{current}} - E$
        \State $C_{\text{max}} \gets C_{\text{max}} \cdot (1 - \eta)$
        \State $\text{DailyActionCount}[d] \gets \text{DailyActionCount}[d] + 1$
    \EndIf
\EndFor
\EndProcedure
\end{algorithmic}
\end{algorithm}

\begin{algorithm}[!ht]
\caption{MILP-Optimised Trading Algorithm with ML-Enhanced Price Forecasts}\label{alg:ml}
\begin{algorithmic}[1]
\Require $C_{\text{init}},~C_{\text{current}},~\eta,~A_{\text{max}},~\Delta t,~P_{\text{charge}},~P_{\text{discharge}},~R$
\LeftComment{ \textit{$C_{\text{init}}$: initial battery capacity (MWh)}}
\LeftComment{ \textit{$C_{\text{current}}$: current SOC (MWh)}}
\LeftComment{ \textit{$\eta$: battery degradation rate}}
\LeftComment{ \textit{$A_{\text{max}}$: max daily actions}}
\LeftComment{ \textit{$\Delta t$: settlement period}}
\LeftComment{ \textit{$P_{\text{charge}}$, $P_{\text{discharge}}$: threshold prices}}
\LeftComment{ \textit{$R$: total revenue}}

\Procedure{MLBidStrategy}{$\text{ForecastPrice}[t],~\text{MarketPrice}[t],~\text{Date}[t]$}
\State Initialise $C_{\text{max}} \gets C_{\text{init}},~C_{\text{current}} \gets 0.5 \cdot C_{\text{init}}$
\State Initialise $\text{DailyActionCount}[\cdot] \gets 0$
\For{$t \in [1,~T]$}
    \State Generate short-term price forecasts with ML model
    \State Compute weighted average price prediction $\hat{p}_t$
    \State Optimise schedule with MILP using $\hat{p}_t$ as input
    \State $d \gets \text{Date}[t]$
    \State $p \gets \text{MarketPrice}[t]$
    \State $decision \gets$ optimiser decision at time $t$: \textbf{Charge}, \textbf{Discharge}, or \textbf{Hold}
    \If{$\text{DailyActionCount}[d] \geq A_{\text{max}}$ \textbf{and} predicted revenue $< \text{threshold}$}
        \State \textbf{continue}
    \EndIf
    \If{$decision = \text{Charge}$ \textbf{and} $P_{\text{charge}} \geq p$ \textbf{and} $C_{\text{current}} < C_{\text{max}}$}
        \State $E \gets \min(P_{\text{max}} \cdot \Delta t,~C_{\text{max}} - C_{\text{current}})$
        \State $R \gets R - E \cdot p$
        \State $C_{\text{current}} \gets C_{\text{current}} + E$
        \State $C_{\text{max}} \gets C_{\text{max}} \cdot (1 - \eta)$
        \State $\text{DailyActionCount}[d] \gets \text{DailyActionCount}[d] + 1$
    \ElsIf{$decision = \text{Discharge}$ \textbf{and} $P_{\text{discharge}} \leq p$ \textbf{and} $C_{\text{current}} > 0$}
        \State $E \gets \min(P_{\text{max}} \cdot \Delta t,~C_{\text{current}})$
        \State $R \gets R + E \cdot p$
        \State $C_{\text{current}} \gets C_{\text{current}} - E$
        \State $C_{\text{max}} \gets C_{\text{max}} \cdot (1 - \eta)$
        \State $\text{DailyActionCount}[d] \gets \text{DailyActionCount}[d] + 1$
    \EndIf
\EndFor
\EndProcedure
\end{algorithmic}
\end{algorithm}






\clearpage



\setlength{\baselineskip}{0pt} 

\renewbibmacro*{urldate}{
(retrieved \printfield{urlday}/\printfield{urlmonth}/\printfield{urlyear})
} 

{\renewcommand*\MakeUppercase[1]{#1}%
\printbibliography[heading=bibintoc,title={\bibtitle}]}

\end{document}